\documentclass[twoside]{article}

\usepackage[dvipsnames]{xcolor}
\usepackage[accepted]{aistats2022}
\usepackage[colorlinks = true,
            linkcolor = black,
            urlcolor  = blue,
            citecolor = blue,
            anchorcolor = blue,
            pagebackref = true]{hyperref}

\usepackage[round]{natbib}

\usepackage[utf8]{inputenc} %
\usepackage[T1]{fontenc}    %
\usepackage{hyperref}       %
\usepackage{url}            %
\usepackage{booktabs}       %
\usepackage{amsfonts}       %
\usepackage{nicefrac}       %
\usepackage{microtype}      %
\usepackage{xcolor}         %
\usepackage{amsmath}
\usepackage{amsthm}
\usepackage{dsfont}
\usepackage{amssymb}
\usepackage{amscd}
\usepackage{mathtools}
\mathtoolsset{showonlyrefs=true}
\usepackage{enumerate}
\usepackage{authblk}
\usepackage{ bbold }
\usepackage{algorithmic}
\usepackage{algorithm}
\usepackage{authblk}
\usepackage{wrapfig}

\newtheorem{theorem}{Theorem}
\newtheorem*{recall_thm1}{Theorem~\ref{thm:equivalence}}
\newtheorem*{recall_thm2}{Theorem~\ref{thm:error}}
\newtheorem{lemma}{Lemma}

\newcommand{\argmax}{\operatorname*{argmax}} 

\newcommand{\sm}{\operatorname*{softmax}}
\newcommand{\kl}[2]{\operatorname*{KL}(#1||#2)}

\newcommand{\hc}{\mathcal{H}}

\newcommand{\states}{\mathcal{S}}
\newcommand{\actions}{\mathcal{A}}
\newcommand{\simplex}{\Delta^\states_\mathcal{A}}

\newcommand{\E}{\mathbb{E}}

\newcommand{\1}{\mathbf{1}}
\DeclareMathOperator{\FC}{FC}

\begin{document}

\runningauthor{Nino Vieillard, Marcin Andrychowicz,  Anton Raichuk, Olivier Pietquin, Matthieu Geist}

\twocolumn[

\aistatstitle{Implicitly Regularized RL with Implicit Q-values}

\aistatsauthor{Nino Vieillard$^{1,2}$ \And Marcin Andrychowicz$^{1}$ \And  Anton Raichuk$^{1}$}
\aistatsauthor{Olivier Pietquin$^{1}$ \And Matthieu Geist$^{1}$}

\aistatsaddress{$^{1}$Google Research, Brain Team \quad $^{2}$Universit\'e de Lorraine, CNRS, Inria, IECL, F-54000 Nancy, France}]

\begin{abstract}

The $Q$-function is a central quantity in many Reinforcement Learning (RL) algorithms for which RL agents behave following a (soft)-greedy policy w.r.t. to $Q$. It is a powerful tool that allows action selection without a model of the environment and even without explicitly modeling the policy. Yet, this scheme can only be used in discrete action tasks, with small numbers of actions, as the softmax over actions cannot be computed exactly otherwise.  More specifically, the usage of function approximation to deal with continuous action spaces in modern actor-critic architectures intrinsically prevents the exact computation of a softmax. We propose to alleviate this issue by parametrizing the $Q$-function \emph{implicitly}, as the sum of a log-policy and a value function. We use the resulting parametrization to derive a practical off-policy deep RL algorithm, suitable for large action spaces, and that enforces the softmax relation between the policy and the $Q$-value. We provide a theoretical analysis of our algorithm: from an Approximate Dynamic Programming perspective, we show its equivalence to a regularized version of value iteration, accounting for both entropy and Kullback-Leibler regularization, and that enjoys beneficial error propagation results.  We then evaluate our algorithm on classic control tasks, where its results compete with state-of-the-art methods.

\end{abstract}

\section{INTRODUCTION}

A large body of reinforcement learning (RL) algorithms, based on approximate dynamic programming (ADP)~\citep{bertsekas1996neuro,scherrer2015approximate}, operate in two steps: a greedy step, where the algorithm learns a policy that maximizes a $Q$-value, and an evaluation step, that (partially) updates the $Q$-values towards the $Q$-values of the policy. A common improvement to these techniques is to use regularization, that prevents the new updated policy from being too different from the previous one, or from a fixed ``prior'' policy. For example, Kullback-Leibler (KL) regularization keeps the policy close to the previous iterate~\citep{vieillard2020leverage}, while entropy regularization keeps the policy close to the uniform one~\citep{haarnoja2018soft}. Entropy regularization, often used in this context~\citep{ziebart2010modeling}, modifies both the greedy step and the evaluation step so that the policy jointly maximizes its expected return and its entropy. In this framework, the solution to the policy optimization step is simply a softmax of the $Q$-values over the actions. In small discrete action spaces, the softmax can be computed exactly: one only needs to define a \textit{critic} algorithm, with a single loss that optimizes a $Q$-value. However, in large multi-dimensional --~or even continuous~-- action spaces, one needs to estimate it. This estimation is usually done by adding an \emph{actor} loss, that optimizes a policy to fit this softmax. It results in an (off-policy) \textit{actor-critic} algorithm, with two losses that are optimized simultaneously\footnote{We refer here specifically to \emph{off-policy} actor-critics, built on a value-iteration-like scheme. For on-policy actor-critics, losses are optimized sequentially, and the policy is usually fully evaluated.} \citep{degris2012off}. This additional optimization step introduces supplementary errors to the ones already created by the approximation in the evaluation step. 

\looseness=-1
To remove these extraneous approximations, we introduce the  Implicit $Q$-values (IQ) algorithm, that deviates from classic actor-critics, as it optimizes a policy and a value in a single loss. The core idea is to implicitly represent the $Q$-value as the sum of a value function and a log-policy. This representation ensures that the policy is the \emph{exact} softmax of the $Q$-value, \emph{despite the use of any approximation scheme}. We use this to design a practical model-free deep RL algorithm  that optimizes with a single loss a policy network and a value network, built on this implicit representation of a $Q$-value. To better understand it, we abstract this algorithm to an ADP scheme, IQ-DP, and use this point of view to provide a detailed theoretical analysis. It relies on a key observation, that shows an equivalence between IQ-DP and a specific form of regularized Value Iteration (VI). This equivalence explains the role of the components of IQ: namely, IQ performs entropy and KL regularization. It also allows us to derive strong performance bounds for IQ-DP. In particular, we show that the errors made when following IQ-DP are compensated along iterations.

Parametrizing the $Q$-value as a sum of a log-policy and a value is reminiscent of the dueling architecture~\citep{wang2016dueling}, that factorizes the $Q$-value as the sum of an advantage and a value. In fact, we show that it is a limiting case of IQ in a discrete actions setting. 
This link highlights the role of our policy, which calls for a discussion on its necessary parametrization. 

Finally, we empirically validate IQ. We evaluate our method on several classic continuous control benchmarks: locomotion tasks from Openai Gym~\citep{brockman2016openai}, and hand manipulation tasks from the Adroit environment~\citep{rajeswaran2017learning}. On these environments, IQ reaches performances competitive with state-of-the-art actor-critic methods. %

\section{IMPLICIT $Q$-VALUE PARAMETRIZATION}

\begin{figure}
\centering
  \includegraphics[trim={0 3cm 0cm 5.5cm},clip,width=.4\textwidth]{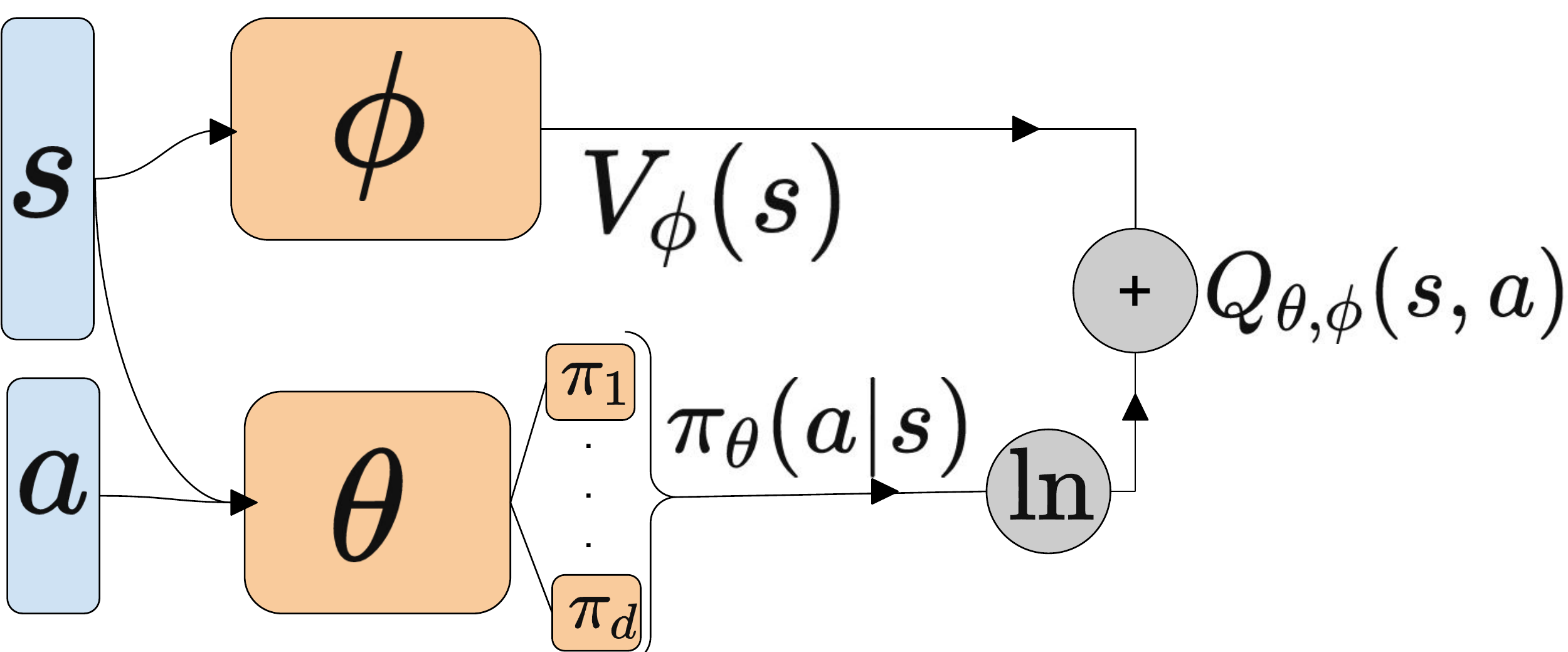}
  \caption{view of the IQ parametrization.}
\label{fig:view}
\end{figure}
\label{sec:IQ}
We consider the standard Reinforcement Learning (RL) setting, formalized as a Markov Decision Process (MDP). An MDP is a tuple $\{\states,\actions,P,r,\gamma\}$. $\states$ and $\actions$ are the finite state and action spaces\footnote{We restrict to finite spaces for the sake of analysis, but our approach applies to continuous spaces.}, $\gamma\in[0,1)$ is the discount factor and $r:{\states\times\actions}\rightarrow [-R_{max}, R_{max}]$ is the bounded reward function. Write $\Delta_X$ the simplex over the finite set $X$. The dynamics of an MDP are defined by a Markovian transition kernel $P\in\Delta_{\states}^{\states\times\actions}$, where $P(s'|s,a)$ is the probability of transitioning to state $s'$ after taking action $a$ in $s$. An RL agent acts through a stationary stochastic policy $\pi\in\Delta_\actions^\states$, a mapping from states to distribution over actions. 
The quality of a policy is quantified by the value function, $V_\pi(s) = \mathbb{E}_\pi[\sum_{t=0}^\infty \gamma^t r(s_t, a_t) | s_0=s]$. The $Q$-function is a useful extension, which notably allows choosing a (soft)-greedy action in a model-free setting, $Q_\pi(s,a) = r(s,a) + \mathbb{E}_{s' | s,a}[V_\pi(s')]$. An optimal policy is one that achieve the highest expected return, $\pi_* = \argmax_\pi V_\pi$.

\looseness=-1
A classic way to design practical algorithms beyond the tabular setting is to adopt the Actor-Critic perspective. In this framework, an RL agent parametrizes a policy $\pi_\theta$ and a $Q$-value $Q_\psi$ with function approximation, usually through the use of neural networks, and aims at estimating an optimal policy. The policy and the $Q$-function are then updated by minimizing two losses: the actor loss corresponds to the greedy step, and the  critic loss to the evaluation step. The weights of the policy and  $Q$-value networks are regularly frozen into \emph{target} weights $\bar{\psi}$ and $\bar{\theta}$. With entropy regularization, the greedy step amounts to finding the policy that maximizes $\E_{s \sim \states, a \sim \pi_\theta}[Q_{\bar\psi}(s,a) - \tau\ln\pi_\theta(a|s)]$ (maximize the $Q$-value with stochastic enough policy). The solution to this problem is simply 
$\pi_\theta(\cdot|s) = \sm({Q_{\bar{\psi}}(s,\cdot)}/{\tau}), \label{eq:softmax}$
which is the result of the greedy step of regularized Value Iteration (VI)~\citep{geist2019theory} and, for example, how the optimization step of Soft Actor-Critic \citep[SAC]{haarnoja2018soft} is built. In a setting where the action space is discrete and small, it amounts to a simple softmax computation. However, on more complex action spaces (continuous, and/or with a higher number of dimensions: as a reference, the Humanoid-v2 environment from Openai Gym~\citep{brockman2016openai} has an action space of dimension $17$), it becomes prohibitive to use the exact solution. In this case, the common practice is to resort to an approximation with a parametric distribution model. In many actor critic algorithms (SAC, MPO~\citep{abdolmaleki2018maximum},~...), the policy is modelled as a Gaussian distribution over actions. It introduces approximation errors, resulting from the partial optimization process of the critic, and inductive bias, as a Gaussian policy cannot represent an arbitrary softmax distribution. We now turn to the description of our core contribution: the Implicit $Q$-value (IQ) algorithm, introduced to mitigate this discrepancy.

IQ implicitly parametrizes a $Q$-value via an explicit parametrization of a policy and a value, as visualized in Fig.~\ref{fig:view}. Precisely, from a policy network $\pi_\theta$ and a value network $V_\phi$, we define our implicit $Q$-value as
\begin{equation}
\label{eq:IQ-param}
    Q_{\theta, \phi}(s,a) = \tau\ln\pi_\theta(a|s) + V_\phi(s).
\end{equation}
Since $\pi_\theta$ is constrained to be a distribution over the actions, we have by construction that $\pi_\theta(a|s) = \sm(Q_{\theta,\phi}/\tau)$, the solution of the regularized greedy step (see Appx.~\ref{subappx:about} for a detailed proof). Hence, the consequence of using such a parametrization is that the greedy step is performed exactly, even in the function approximation regime. Compared to the classic actor-critic setting, it thus gets rid of the errors created by the actor. Note that calling $V_\phi$ a value makes sense, since following the same reasoning we have that $V_\phi(s) = \tau\ln\sum_{a'}\exp (Q_{\theta, \phi}(s,a')/\tau)$, a soft version of the value. With this parametrization in mind, one could derive a deep RL algorithm from any value-based loss using entropy regularization.  We conserve the fixed-point approach of the standard actor-critic framework, $\theta$ and $\phi$ are regularly copied to $\bar{\theta}$ and $\bar{\phi}$, and we design an off-policy algorithm, working on a replay buffer of transitions $(s_t, a_t, r_t, s_{t+1})$ collected during training. Consider two hyperparameters, $\tau\in(0, \infty)$ and $\alpha\in(0,1)$ that we will show in Sec.~\ref{sec:analysis} control two forms of regularization.
The policy and value are optimized jointly by minimizing the loss
\begin{multline}
\label{eq:IQ:loss}%
    \mathcal{L}_{\text{IQ}}(\theta,\phi) = \hat{\E}\Big[\big(r_t + \alpha\tau\ln\pi_{\bar{\theta}}(a_t|s_t) + \gamma V_{\bar{\phi}}(s_{t+1}) \\
    - \tau\ln\pi_\theta(a_t|s_t) - V_\phi(s_t)\big)^2\Big],
\end{multline}
\looseness=-1
where $\hat{\E}$ denote the empirical expected value over a dataset of transitions. IQ consists then in a single loss that optimizes jointly a policy and a value. This brings a notable remark on the role of $Q$-functions in RL. Indeed, $Q$-learning was introduced by~\citet{watkins1992q} -- among other reasons -- to make greediness possible without a model (using a value only, one needs to maximize it over all possible successive states, which requires knowing the transition model), and consequently derive practical, model-free RL algorithms. Here however, IQ illustrates how, with the help of regularization, one can derive a model-free algorithm that does not rely on an explicit $Q$-value.

\section{ANALYSIS}
\label{sec:analysis}
In this section, we explain the workings of the IQ algorithm defined by Eq.~\eqref{eq:IQ:loss} and detail the influence of its hyperparameters. We abstract IQ into an ADP framework, and show that, from that perspective, it is equivalent to a Mirror Descent VI (MD-VI) scheme~\citep{geist2019theory}, with both entropy and KL regularization. Let us first introduce some useful notations. We make use of the actions partial dot-product notation: for $u,v\in\mathbb{R^{\states\times\actions}}$, we define $\langle u, v\rangle = \big(\sum_{a\in\actions}u(s,a)v(s,a)\big)_s \in \mathbb{R}^\states$. For any $V\in\mathbb{R}^\states$, we have for any $(s,a)\in\states\times\actions$ $PV(s,a) = \sum_{s'}P(s'|s,a)V(s')$. We will define regularized algorithms, using the entropy of a policy, $\mathcal{H}(\pi) = -\langle \pi, \ln\pi\rangle$, and the KL divergence between two policies, $\kl{\pi}{\mu}=\langle\pi, \ln\pi - \ln\mu\rangle$. The $Q$-value of a policy is the unique fixed point of its Bellman operator $T_\pi$ defined for any $Q\in\mathbb{R}^{\states\times\actions}$ as $T_\pi Q = r + \gamma P\langle \pi,Q \rangle$. We denote $Q_*=Q_{\pi_*}$ the optimal $Q$-value (the $Q$-value of the optimal policy). When the MDP is entropy-regularized with a temperature $\tau$, a policy  $\pi$ admits a \textit{regularized} $Q$-value $Q_\pi^\tau$, the fixed point of the regularized Bellman operator $T_\pi^\tau Q = r + \gamma P\langle\pi, Q - \tau\ln\pi\rangle$. A regularized MDP admits an optimal \textit{regularized} policy $\pi_*^{\tau}$ and a unique optimal \textit{regularized} $Q$-value $Q_*^\tau$~\citep{geist2019theory}.

\subsection{Ideal case}
\looseness=-1
First, let us look at the ideal case, \emph{i.e.} when $\mathcal{L}_{\text{IQ}}$ is exactly minimized at each iteration (tabular representation, dataset covering the whole state-action space, expectation rather than sampling for transitions). In this context, IQ can be understood as a Dynamic Programming (DP) scheme that iterates on a policy $\pi_{k+1}$ and a value $V_k$. They are respectively equivalent to the target networks $\pi_{\bar\theta}$ and $V_{\bar\phi}$, while the next iterate $(\pi_{k+2}, V_{k+1})$ matches the solution $(\pi_\theta, V_\phi)$ of the optimization problem in Eq.~\eqref{eq:IQ:loss}. We call the scheme IQ-DP$(\alpha, \tau)$ and one iteration is defined by choosing $(\pi_{k+2}, V_{k+1})$ such that the squared term in Eq.~\eqref{eq:IQ:loss} is $0$, %
\begin{equation}
\label{eq:IQ-DP}
    \tau\ln \pi_{k+2} + V_{k+1} = r + \alpha\tau\ln\pi_{k+1} + \gamma P V_k. 
\end{equation}
This equation is well-defined, due to the underlying constraint that $\pi_{k+2} \in \simplex$ (the policy must be a distribution over actions), that is $\sum_{a\in\actions}\pi(a|s) = 1$ for all $s\in\states$.  The basis for our discussion will be the equivalence of this scheme to a version of regularized VI. Indeed, we have the following result, proved in Appendix~\ref{appx:adp}.
\begin{theorem}
\label{thm:equivalence}
For any $k\geq1$, let $(\pi_{k+2}, V_{k+1})$ be the solution of IQ-DP$(\alpha, \tau)$ at step $k$. We have that
\begin{equation}
\begin{cases}
    \pi_{k+2} = \argmax\langle\pi, r + \gamma P V_k\rangle + (1 - \alpha)\tau \hc(\pi) \\
              \hfill - \alpha\tau \kl{\pi}{\pi_{k+1}} \\
    V_{k+1} = \langle\pi_{k+2}, r + \gamma P V_k\rangle + (1 - \alpha)\tau \hc(\pi_{k+2}) \\
                \hfill - \alpha\tau \kl{\pi_{k+2}}{\pi_{k+1}}
\end{cases} 
\end{equation}
so IQ-DP($\alpha, \tau$) produces the same sequence of policies as a value-based version of Mirror Descent VI, MD-VI$(\alpha\tau, (1-\alpha)\tau)$~\citep{vieillard2020leverage}.
\end{theorem}

\paragraph{Discussion.} The previous results shed a first light on the nature of the IQ method. Essentially, IQ-DP is a parametrization of a VI scheme regularized with both  entropy and KL divergence, MD-VI$(\alpha\tau, (1-\alpha)\tau)$. This first highlights the role of the hyperparameters, as its shows the interaction between the two forms of regularization. The value of $\alpha$ balances between those two: with $\alpha=0$, IQ-DP reduces to a classic VI regularized with entropy; with $\alpha=1$ only the KL regularization will be taken into account. The value of $\tau$ then controls the amplitude of this regularization. In particular, in the limit $\alpha=0, \tau \rightarrow 0$, we recover the standard VI algorithm. This results also justifies the soundness of IQ-DP. Indeed, this MD-VI scheme is known to converge to $\pi_*^{(1-\alpha)\tau}$ the optimal policy of the regularized MDP~\citep[Thm.~2]{vieillard2020leverage} and this results readily applies to IQ\footnote{\citet{vieillard2020leverage} show this for $Q$-functions, but it can straightforwardly be extended to value functions.}. Another consequence is that it links IQ to  Advantage Learning (AL)~\citep{bellemare2016increasing}. Indeed, AL is a limiting case of MD-VI when $\alpha>0$ and $\tau\rightarrow0$ \citep{vieillard2020munchausen}. Therefore, IQ also generalizes AL, and the $\alpha$ parameter can be interpreted as the advantage coefficient. Finally, a key observation is that IQ performs KL regularization implicitly, the way it was introduced by Munchausen RL~\citep{vieillard2020munchausen}, by augmenting the reward with the $\alpha\tau\ln\pi_{k+1}$ term (Eq.~\eqref{eq:IQ-DP}). This observation will have implications discussed next. %

\subsection{Error propagation result}
\label{subsec:error}
Now, we are interested in understanding how errors introduced by the function approximation used propagate along iterations. At iteration $k$ of IQ, denote $\pi_{k+1}$ and $V_k$  the target networks. In the approximate setting, we do not solve Eq.~\eqref{eq:IQ-DP}, but instead, we minimize $\mathcal{L}(\theta, \phi)$ with stochastic gradient descent. This means that $\pi_{k+2}$ and $V_{k+1}$ are the result of this optimization, and thus the next target networks. The optimization process introduces errors, that come from many sources: partial optimization, function approximation (policy and value are approximated with neural networks), finite data, etc. We study the impact of these errors on the distance between the optimal $Q$-value of the MDP and the regularized $Q$-value of the current policy used by IQ, $Q_{\pi_{k+1}}^{(1-\alpha)\tau}$. We insist right away that $Q_{\pi_{k+1}}^{(1-\alpha)\tau}$ is not the learned, implicit $Q$-value, but the actual $Q$-value of the policy computed by IQ in the regularized MDP. We have the following result concerning the error propagation.
\begin{theorem} \label{thm:error}
Write $\pi_{k+1}$ and $V_k$ the $k^{th}$ update of respectively the target policy and value networks. Consider the error at step $k$, $\epsilon_k\in\mathbb{R}^{\states\times\actions}$, as the difference between the ideal and the actual updates of IQ. Formally, we define the error as, for all $k \geq 1$,
$$
    \epsilon_k = \tau\ln\pi_{k+2} + V_{k+1} - (r + \alpha\tau\ln\pi_{k+1} + \gamma P V_k),
$$
and the moving average of the errors as
$
    E_k = (1 - \alpha) \sum_{j=1}^{k}\alpha^{k-j}\epsilon_j 
$.
We have the following results for two different cases depending on the value of $\alpha$. Note that when $\alpha<1$, we bound the distance to regularized optimal $Q$-value.
\begin{enumerate}
    \item General case: $0 < \alpha <1$ and $\tau> 0$, entropy and KL regularization together:
\begin{multline}
      \|Q_*^{(1-\alpha)\tau} - Q_{\pi_k}^{(1-\alpha)\tau}\|_{\infty} \leq\\
       \frac{2}{(1 - \gamma)^2}\left((1-\gamma)\sum_{j=1}^k \gamma^{k-j} \Vert E_j \Vert_\infty \right) + o\left(\frac{1}{k}\right).
\end{multline}
     \item Specific case $\alpha=1$, $\tau > 0$, use of KL regularization alone:
\begin{equation}
     \|Q_* - Q_{\pi_k}\|_{\infty}\leq 
    \frac{2}{1-\gamma} \left\|\frac{1}{k}\sum_{j=1}^k \epsilon_j\right\|_\infty
    + O\left(\frac{1}{k}\right). %
\end{equation}
\end{enumerate}
\end{theorem}
\begin{proof}[Sketch of proof.]
The full proof is provided in Appendix~\ref{appx:error}. We build upon the connection we established between IQ-DP and a VI scheme regularized by both KL and entropy in Thm.~\ref{thm:equivalence}. By injecting the proposed representation into the classic MD-VI scheme, we can build upon the analysis of~\citet[Thm.~1 and~2]{vieillard2020leverage} to provide these results.
\end{proof}

\paragraph{Impact of KL regularization.} The KL regularization term, and specifically in the MD-VI framework, is discussed extensively by~\citet{vieillard2020leverage}, and we refer to them for in-depth analysis of the subject. We recall here the main interests of KL regularization, as illustrated by the bounds of Thm~\ref{thm:error}. In the second case, where it is the clearest (only KL is used), we observe a beneficial property of KL regularization: Averaging of errors. Indeed, in a classic non-regularized VI scheme
~\citep{scherrer2015approximate}, the error $ \|Q_* - Q_{\pi_{\theta}}\|$ would depend on a moving average of the norms of the errors $(1-\gamma)\sum_{j=1}^{k}\gamma^{k-j}\Vert\epsilon_k\Vert_\infty$, while with the KL it depends on the norm of the average of the errors $(1/k)\Vert\sum_{j=1}^{k}\epsilon_k\Vert$. In a simplified case where the errors would be i.i.d. and zero mean, this would allow convergence of approximate MD-VI, but not of approximate VI. In the case $\alpha<1$, where we introduce entropy regularization, the impact is less obvious, but we still transform a sum of norm of errors into a sum of moving average of errors, which can help by reducing the underlying variance.

\paragraph{Link to Munchausen RL.} As stated in the sketched proof, Thm.~\ref{thm:error} is a consequence of~\citep[Thm.~1 and~2]{vieillard2020leverage}. A crucial limitation of this work is that the analysis only applies when no errors are made in the greedy step. This is possible in a relatively simple setting, with tabular representation, or with a linear parametrization of the $Q$-function. However, in the general case with function approximation, exactly solving the optimization problem regularized by KL is not immediately possible: the solution of the greedy step of MD-VI$(\alpha\tau, (1-\alpha)\tau)$ is $\pi_{k+2} \propto \exp(Q_{k+1}/\tau)\pi_k^\alpha$ (where $Q_{k+1} = r + \gamma P V_k$), so computing it exactly would require remembering every $\pi_j$ during the procedure, which is not feasible in practice. A workaround to this issue was introduced by~\citet{vieillard2020munchausen} as Munchausen RL: the idea is to augment the reward by the log-policy, to implicitly define a KL regularization term, while reducing the greedy step to a softmax. As mentioned before, in small discrete action spaces, this allows to compute the greedy step exactly, but it is not the case in multidimensional or continuous action spaces, and thus Munchausen RL loses its interest in such domains. With IQ, we utilize the Munchausen idea to implicitly define the KL regularization; but with our parametrization, the exactness of the greedy step holds even for complex action spaces: recall that the parametrization defined in Eq.~\eqref{eq:IQ-param} enforces that the policy is a softmax of the (implicit) $Q$-value. Thus, IQ can be seen as an extension of Munchausen RL to multidimensional and continuous action spaces.

\subsection{Link to the dueling architecture}

Dueling Networks (DN) were introduced as a variation of the seminal Deep Q-Networks (DQN, \citet{mnih2015human}), and have been empirically proven to be efficient (for example by~\citet{hessel2018rainbow}). The idea is to represent the $Q$-value as the sum of a value and an advantage. In this setting, we work with a notion of advantage defined over $Q$-functions (as opposed to defining the advantage as a function of a policy). For any $Q\in\mathbb{R^{\states\times\actions}}$, its advantage $A_Q$ is  defined as $A_Q(s,a) = Q(s,a) - \max_{a'\in\actions}Q(s,a')$. The advantage encodes a sub-optimality constraint: it has negative values and its maximum over actions is $0$. \citet{wang2016dueling} propose to learn a $Q$-value by defining an advantage network $F_\Theta$ and a value network $V_\Phi$, which in turn define a $Q$-value $Q_{\Theta, \Phi}$ as
\begin{equation}
\label{eq:advantage}
    Q_{\Theta, \Phi}(s,a) =  \textcolor{BrickRed}{F_\Theta(s,a) - \max_{a'\in\actions}F_\Theta(s,a')} + V_\Phi(s).
\end{equation}
Subtracting the maximum over the actions ensures that $F_\Theta$ indeed represents an advantage (in \textcolor{BrickRed}{red}). Note that dueling DQN was designed for discrete settings, where computing the maximum over actions is not an issue.

In IQ, we need a policy network that represents a distribution over the actions. There are several practical ways to represent the policy, discussed in Sec~\ref{sec:practical}. For the sake of simplicity, let us for now assume that we are in a mono-dimensional discrete action space, and that we use a common scaled softmax representation. Specifically, our policy is represented by a neural network (eg. fully connected) $F_\theta$, that maps state-action pairs to logits $F_\theta(s,a)$. The policy is then defined as $\pi_\theta(\cdot|s) = \sm(F_\theta(s,\cdot)/\tau)$. Directly from the definition of the softmax, we observe that $\tau\ln\pi_\theta(a|s) = F_\theta(s,a) - \tau\ln\sum_{a'\in\actions}\exp(F_\theta(s,a')/\tau)$. The second term is a classic scaled logsumexp over the actions, a soft version of the maximum: when $\tau\rightarrow 0$, we have that $\tau\ln\sum_a'\exp(F(s,a')/\tau) \rightarrow \max_{a'}F(s,a')$. Within the IQ parametrization, we have
\begin{equation}
\label{eq:soft-advantage}
    Q_{\theta, \phi}(s,a) =  \textcolor{BrickRed}{F_\theta(s,a) - \tau\ln\sum_{a'\in\actions}\exp\frac{F_\theta(s,a')}{\tau}} + V_\phi(s),
\end{equation}
\looseness=-1
where we highlighted in \textcolor{BrickRed}{red} the soft advantage component, analog to the advantage in Eq.~\eqref{eq:advantage}, which makes a clear link between IQ and DN. In this case (scaled softmax representation), the IQ parametrization generalizes the dueling architecture, retrieved when $\tau\rightarrow0$ (and with an additional AL term whenever $\alpha>0$, see Sec.~\ref{sec:analysis}). In practice,~\citet{wang2016dueling} use a different parametrization of the advantage, replacing the maximum by a mean, defining $Q_{\Theta, \Phi}(s,a) =  A_\Theta(s,a) - |\actions|^{-1}\sum_{a'\in\actions}A_\Theta(s,a') + V_\Phi(s)$. We could use a similar trick and replace the logsumexp by a mean in our policy parametrization, but in our case this did not prove to be efficient in practice.

We showed how the log-policy represents a soft advantage. While this explicits  its role in the learning procedure, it also raises questions about which representation would be the most suited for optimization.

\section{PRACTICAL CONSIDERATIONS} \label{sec:practical}
We now describe key practical issues encountered when choosing a policy representation. The main one comes from the delegation of the representation power of the algorithm to the policy network. In a standard actor-critic algorithm -- take SAC for example, where the policy is parametrized as a Gaussian distribution -- the goal of the policy is mainly to track the maximizing action of the $Q$-value. Thus, estimation errors can cause the policy to  choose sub-optimal actions, but the inductive bias caused by the Gaussian representation may not be a huge issue in practice, as long as the mean of the Gaussian policy is not too far from the maximizing action. In other words, the representation capacity of an algorithm such as SAC lies mainly in the representation capacity of its $Q$-network.  In IQ, we have a parametrization of the policy that enforces it to be a softmax of an implicit $Q$-value. By doing this, we trade in estimation error -- our greedy step is exact by construction -- for representation power. Indeed, as our $Q$-value is not parametrized explicitly, but through the policy, the representation power of IQ is in its policy network, and a ``simple'' representation might not be enough anymore. For example, if we parameterized the policy as a Gaussian, this would amount to parametrize an advantage as a quadratic function of the action: this would drastically limit what the IQ policy could represent (in terms of soft-advantage). 

\paragraph{Multicategorical policies.} To address this issue, we turn to other, richer, distribution representations. In practice, we consider a multi-categorical discrete softmax distribution. Precisely, we are in the context of a multi-dimensional action space $\actions$ of dimension $d$, each dimension being a bounded interval. %
We discretize each dimension of the space uniformly  in $n$ values $\delta_j$, for $0\leq j\leq n-1$.
It effectively defines a discrete action space $\actions' = \bigtimes_{j=1}^d \actions_j$, with $\actions_j = \{\delta_0, \hdots \delta_{n-1}\}$. A multidimensional action is a vector $a\in\actions'$, and we denote $a^j$ the $j^{\text{th}}$ component of the action $a$. 
Assuming independence between actions conditioned on states, 
a policy $\pi_\theta$ can be factorized as the product of $d$ marginal mono-dimensional policies
$\pi_\theta(a|s) = \prod_{j=1}^d \pi^{j}_\theta(a^j|s)$.
We represent each policy as the softmax of the output of a neural network $F_\theta^j$, an thus we get the full representation
$
\label{eq:multicat}
    \pi_\theta(a|s) =  \prod_{j=1}^d \sm(F_\theta^j(\cdot|s))(a^j).
$
The  $F_\theta^j$ functions can be represented as neural networks with a shared core, which only differ in the last layer. This type of multicategorical policy can represent any distribution (with $n$ high enough) that does not encompass a dependency between the dimensions. The independence assumption is quite strong, and does not hold in general. From an advantage point of view, it assumes that the soft-advantage (\textit{i.e.} the log-policy) can be linearly decomposed along the actions. While this limits the advantage representation, it is a much weaker constraint than paramterizing the advantage as a quadratic function of the action (which would be the case with a Gaussian policy). In practice, these types of policies have been experimented~\citep{akkaya2019solving, tang2020discretizing}, and have proven to be efficient on continuous control tasks.

\looseness=-1
Even richer policy classes can be explored. To account for dependency between  dimensions, one could envision auto-regressive multicategorical representations, used for example to parametrize a $Q$-value by~\citet{metz2017discrete}. Another approach is to use richer continuous distributions, such as  normalizing flows~\citep{rezende2015variational, ward2019improving}. In this work, we restrict ourselves to the multicategorical setting, which is sufficient to get satisfying results (Sec.~\ref{subsec:results}), and we leave the other options for future work.

\section{RELATED WORK}
\label{sec:related}
\paragraph{Similar parametrizations.} Other algorithms use similar parametrization. First, Path Consistency Learning (PCL,~\citep{nachum2017bridging}) also parametrize the $Q$-value as a sum of a log-policy and a value. Trust-PCL~\citep{nachum2017trust}, builds on PCL by adding a trust region constraint on the policy update, similar to our KL regularization term. A key difference with IQ is that (Trust-)PCL is a residual algorithm, while IQ works around a fixed-point scheme. Shortly, Trust-PCL can be seen as a version of IQ without the target value network $V_{\bar\phi}$. These entropy-regularized residual approaches are derived from the softmax temporal  consistency principle, which allows to consider extensions to  a specific form of multi-step learning (strongly relying on the residual aspect), but they also come with drawbacks, such as introducing a bias in the optimization when the environment is stochastic~\citep{geist2016bellman}.
\citet{dai2018sbeed} proposed SBEED (Smooth Bellmann Error Embedding) to address this bias by replacing the residual minimization with a saddle-point problem. They provide an unbiased algorithm, but that ultimately requires to solve a min-max optimization, more complex than what we propose.
Second, Quinoa~\citep{degrave2019quinoa}  uses a similar loss to Trust-PCL and IQ (without reference to the former Trust-PCL), but does not propose any analysis, and is evaluated only on a few tasks.
Third, Normalized Advantage Function (NAF,~\citet{gu2016continuous}) is designed with similar principles. In NAF, a $Q$-value is parametrized as a value and and an advantage, the latter being quadratic on the action. It matches the special case of IQ with a Gaussian policy, where we recover this quadratic parametrization.
Finally, the action branching architecture~\citep{tavakoli2018action} was proposed to emulate Dueling Networks in the continuous case, by using a similar $Q$-value parametrization with a multicategorical policy representation. This architecture is recovered by IQ in the case $\alpha=0$ and $\tau \rightarrow 0$.
\paragraph{Regularization.} Entropy and KL regularization are used by many other RL algorithms. Notably, from a dynamic programming perspective, IQ-DP(0, $\tau$) performs the same update as SAC -- an entropy regularized VI. This equivalence is however not true in the function approximation regime. Due to the empirical success of SAC and its link to IQ, it will be used as a baseline on continuous control tasks. Other algorithms also use KL regularization, \emph{e.g.} Maximum a posteriori Policy Optimization (MPO,~\citet{abdolmaleki2018maximum}). We refer to~\citet{vieillard2020leverage} for an exhaustive review of algorithms encompassed within the MD-VI framework.

\section{EXPERIMENTS}
\looseness=-1

\paragraph{Environments and metrics.} We evaluate IQ first on the OpenAI Gym environment~\citep{brockman2016openai}. It consists of $5$ locomotion tasks, with action spaces ranging from $3$ (Hopper-v2) to $17$ dimensions (Humanoid-v2). We use a rather long time horizon setting, evaluating our algorithm on $20$M steps on each environments. We also provide result on the Adroit manipulation dataset~\citep{rajeswaran2017learning}, with a similar setting of $20$M environment steps. Adroit is a collection of $4$ hand manipulation tasks, often used in an offline RL setting, but here we use it only as a direct RL benchmark. Out of these $4$ tasks, we only consider $3$ of them: we could not find any working algorithm (baseline or new) on the ``relocate'' task. To summarize the performance of an algorithm, we report the baseline-normalized score along iterations: It normalizes the score so that $0\%$ corresponds to a random score, and $100\%$ to a given baseline. It is defined for one task as
$\text{score} = \frac{\text{score}_\text{algorithm} - \text{score}_\text{random}}{\text{score}_\text{baseline} - \text{score}_\text{random}}$,
where the baseline is the best version of SAC on Mujoco and Adroit after $20$M steps. We report aggregated results, showing the mean and median of these normalized scores along the tasks. Each score is reported as the average over $20$ random seeds. For each experiment, the corresponding standard deviation is reported in Appx.~\ref{subappx:results}.

\looseness=-1
\paragraph{IQ algorithms.} We implement IQ with the Acme~\citep{hoffman2020acme} codebase. It defines two deep neural networks, a policy network $\pi_\theta$ and a value network $V_\phi$. IQ interacts with the environment through $\pi_\theta$, and collect transitions that are stored in a FIFO replay buffer. At each interaction, IQ updates $\theta$ and $\phi$ by performing a step of stochastic gradient descent with  Adam~\citep{kingma2014adam} on $\mathcal{L}_{\text{IQ}}$ (Eq.~\eqref{eq:IQ:loss}). During each step, IQ updates a copy of the weights $\theta$, $\bar\theta$, with a smooth update $\bar\theta \leftarrow (1-\lambda)\bar\theta + \lambda\theta$, with $\lambda\in(0,1)$. It tracks a similar copy $\bar\phi$ of $\phi$. We keep almost all common hyperparameters (networks architecture, $\lambda$, etc.) the same as our main baseline, SAC. We only adjust the learning rate for two tasks, Humanoid and Walker, where we used a lower value: we found that IQ benefits from this, while for SAC we did not observe any improvement (we provide more details and complete results in Appx.~\ref{subappx:results}). Our value network has the same architecture as the SAC $Q$-networks except that the input size is only the state size (as it does not depend on the action). The policy network has the same architecture as the SAC policy network, and differs only by its output: IQ policy outputs a multicategorical policy (so $n\cdot d$ values, where $d$ is the dimensionality of the action space and $n$ is the number of discrete action on each dimension), while SAC policy outputs $2$ $d$-dimensional vectors (mean and diagonal covariance matrix of a Gaussian). We use $n=11$ in our experiments. IQ relies on two other hyperparameters, $\alpha$ and $\tau$. We selected a value of $\tau$ per task suite: $10^{-2}$ on Gym tasks and $10^{-3}$ on Adroit. To make the distinction between the cases when $\alpha=0$ and $\alpha>0$, we denote IQ($\alpha>0$) as M-IQ, for Munchausen-IQ, since it makes use of the Munchausen regularization term. For M-IQ, we found $\alpha=0.9$ to be the best performing value, which is consistent with the findings of~\citet{vieillard2020munchausen}. We report a empirical study of these parameters in the next part of this Section.
Extended explanations are provided in Appx.~\ref{subappx:pc}.

\begin{figure}
    \centering
        \includegraphics[width=\linewidth]{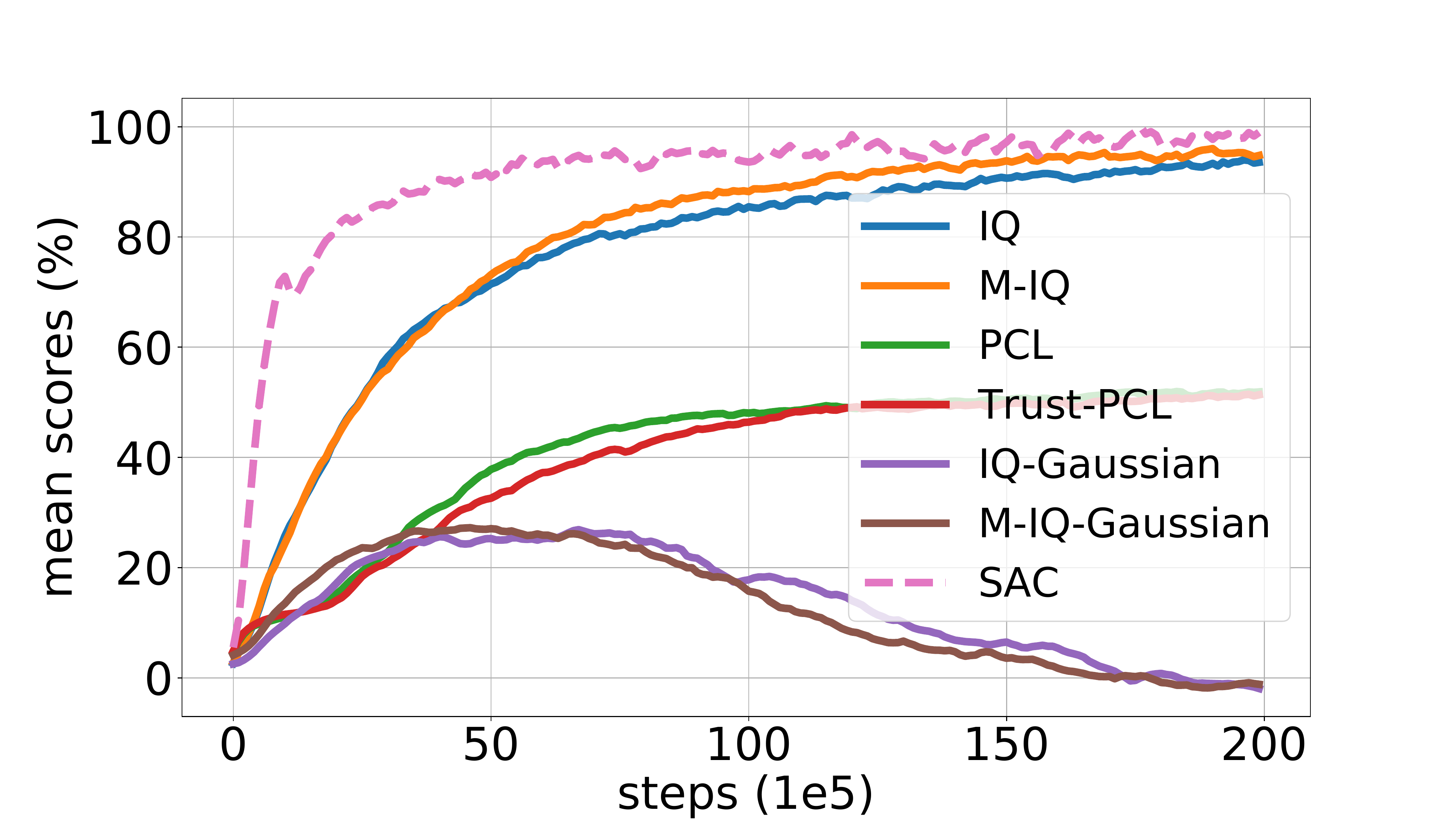}
        \includegraphics[width=\linewidth]{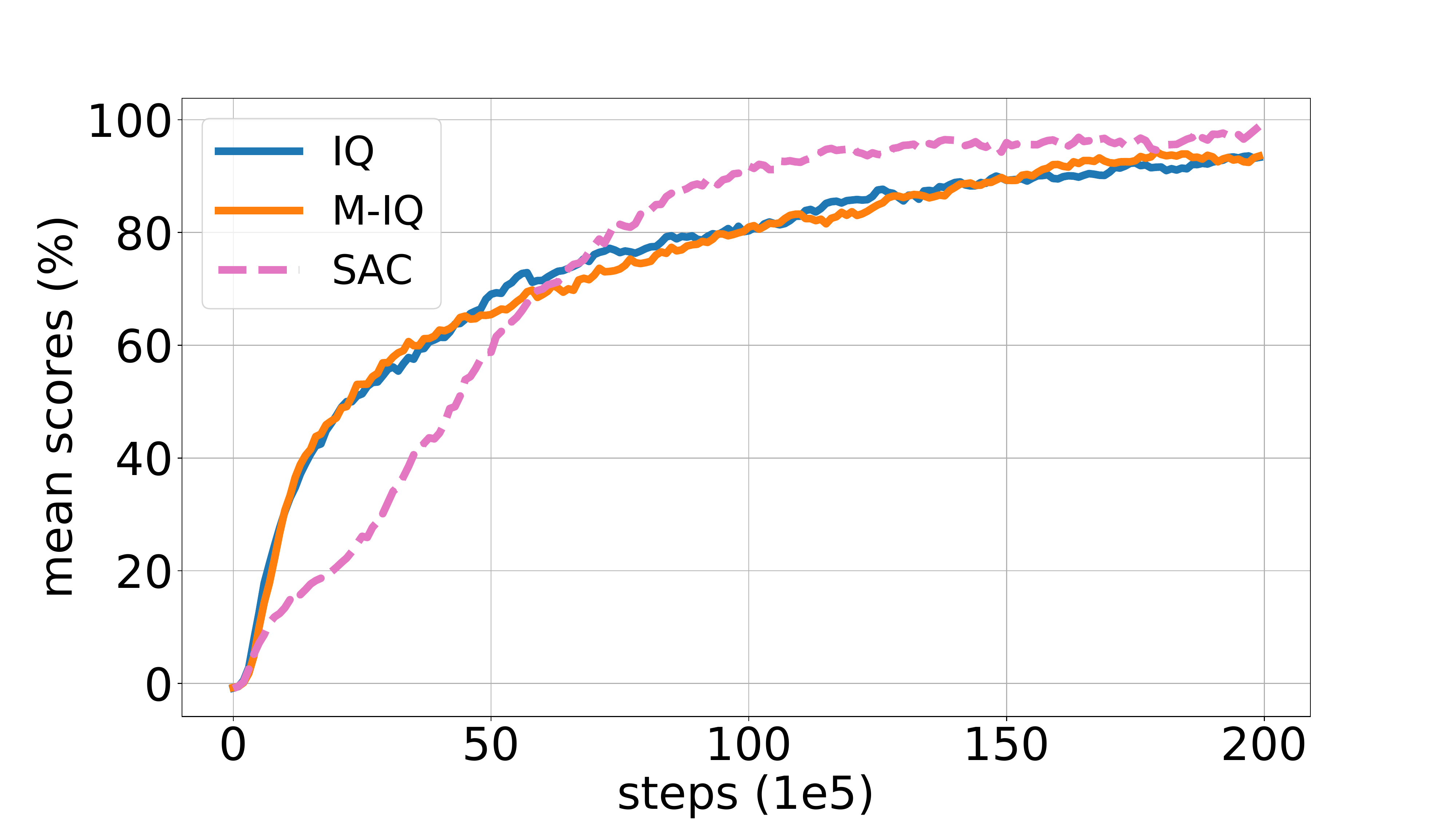}
\vspace{-15pt}
     \caption{SAC-normalized mean scores on Gym (\textbf{top}) and Adroit (\textbf{bottom}).}
    \label{fig:mujoco_iq}
    \label{fig:ablation}
    \label{fig:adroit_iq}
\vspace{-8pt}
\end{figure}

\looseness=-1
\label{subsec:results}
\label{sec:results}
\begin{figure*}
    \centering
    \includegraphics[width=\linewidth]{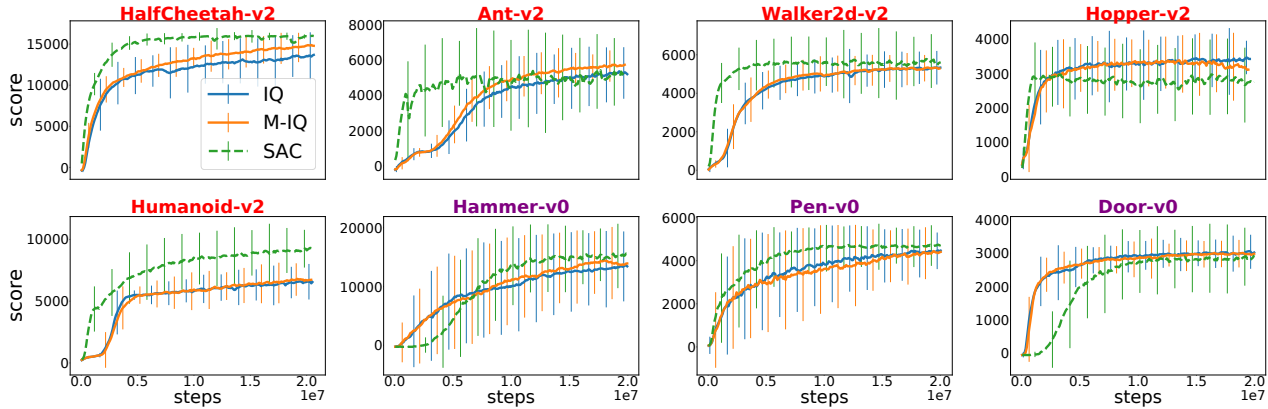}
    \vspace{-20pt}
        \caption{Scores on \textbf{\textcolor{Red}{Gym}} and \textbf{\textcolor{Plum}{Adroit}}. Vertical bars denote +/- empirical standard deviation over $20$ seeds.}
    \label{fig:mujoco_temp}
\end{figure*}

\paragraph{Baselines.} On continuous control tasks, our main baseline is SAC, as it reaches state-of-the-art performance on Mujoco tasks. We compare to the version of SAC that uses an adaptive temperature for reference, but note that for IQ we keep a fixed temperature ($\tau$) setting. To reach its best performance, SAC either uses a specific temperature value per task, or an adaptive scheme that controls the entropy of the policy. This method could be extended to multicategorical policies, but we leave this for future work, and for IQ we use the same value of $\tau$ for all tasks of an environment. We use SAC with the default parameters from~\citet{haarnoja2018soft2} on Gym, and a specifically tuned version of SAC on Adroit. Remarkably, SAC and IQ work with similar hyperparameter ranges on both benchmarks. We only found that using a learning rate of $3\cdot10^{-5}$ (instead of $3\cdot10^{-4}$) gave better performance on Adroit. We also compare IQ to Trust-PCL. It is the closest algorithm to IQ, with a similar parametrization (SBEED also has this parametrization, but all the environments we consider here have detreministic dynamics, and thus there is no bias issue, see Sec.~\ref{sec:related}). To be fair, we compare to our version of Trust-PCL, which is essentially a residual version of IQ, where the target value network $V_{\bar\phi}$ is replaced by the online one. We use Trust-PCL with a fixed temperature, and we tuned this temperature to the environment. We found that Trust-PCL reaches its best performance with significantly lower values of $\tau$ compared to IQ. In the ablation (Fig.~\ref{fig:ablation}) we used $\tau=10^{-4}$ for PCL and Trust-PCL.

\paragraph{Comparison to baselines.} We report aggregated results of IQ and M-IQ on Gym and Adroit in Fig.~\ref{fig:adroit_iq} (median scores can be found in Appx.~\ref{subappx:results}). IQ reaches competitive performance to SAC. It is less sample efficient on Gym (SAC reaches higher performance sooner), but faster on Adroit, and IQ reaches a close final performance on both environments. Detailed scores in Fig.~\ref{fig:mujoco_temp} show how the performance varies across environments. Specifically, IQ outperforms SAC on $3$ of the $8$ considered environments: Ant, Hopper, and Door. On the $5$ others, SAC performs better, but IQ and M-IQ still reach a reasonable performance. Moreover, on almost all environments, all mean performances are within the same confidence interval (Humanoid, and to a less extent HalfCheetah, being notable exceptions, in favor of SAC). The difference in performance between IQ and M-IQ represents the impact of the Munchausen term (\textit{i.e} KL regularization). It is never detrimental, and can even bring some improvement on Gym, but is clearly less useful than in discrete action. Indeed, M-DQN~\citep{vieillard2020munchausen} is designed with the same $\alpha$ parameter but is tailored for discrete actions, and empirically clearly benefits from using $\alpha>0$. We conjecture that this discrepancy comes from the inductive bias we introduce with the multicategorical policy, that could lessen the effect of the Munchausen trick.

\paragraph{Influence of the hyperparameters.} (M)-IQ relies on three key hyperparameters: the temperature $\tau$, the Munchausen coefficient $\alpha$, and the number of bins $n$. The influence of $\alpha$ can be understood as the difference between IQ and M-IQ, discussed above. For the two others parameters, we study their influence with two sweeps, for which we report learning curves in Appx~\ref{subappx:results}. IQ seems to be robust enough to the number of bins, and $n$ turns out to be rather easy to tune: values of $n$ between $7$ and $15$ give similar results. On the other hand, we find that  $\tau$ needs to be selected carefully: while it helps learning, too high values of $\tau$ can be detrimental to the performance, and it highlights that its optimal value might be dependant on the task. In the end, $\tau$ has more influence on IQ than $\alpha$ or $n$, and stands out to be the key element to tune within IQ; thus, adapting the automatic scheduling of the temperature from SAC looks like a promising research direction.

\looseness=-1
\paragraph{Ablation study.} We perform an ablation on important components of IQ in Fig.~\ref{fig:ablation}. \textbf{(1)} We replace the target network by its online counterpart in Eq.~\eqref{eq:IQ:loss}, which gives us Trust-PCL (PCL is obtained by setting $\alpha=0$), a residual version of our method. IQ and M-IQ both outperform Trust-PCL and PCL on Mujoco. \textbf{(2)} We use a Gaussian parametrization of the policy instead of a multicategorical distribution. We observe on Fig.~\ref{fig:ablation} that this causes the performance to drop drastically. This validates the considerations about the necessary complexity of the policy from Section~\ref{sec:practical}.

\section{CONCLUSION}
\looseness=-1
We introduced IQ, a parametrization of a $Q$-value that mechanically preserves the softmax relation between a policy and an implicit $Q$-function. Building on this parametrization, we derived an off-policy algorithm, that learns a policy and a value by minimizing a single loss. We provided insightful analysis that justifies our algorithm and creates meaningful links with the literature (notably with the dueling networks architecture). Specifically, IQ performs entropy and implicit KL regularization on the policy. This kind of regularization was already used and analyzed in RL, but was limited by the difficulty of estimating the softmax of $Q$-function in continuous action settings. IQ ends this limitation by avoiding any approximation in this softmax, effectively extending the analysis of this regularization. This parametrization comes at a cost: it shifts the representation capacity from the $Q$-network to the policy, which makes the use of Gaussian representation ineffective. We solved this issue by considering multicategorical policies, which allowed IQ to reach performance comparable to state-of-the-art methods on classic continuous control benchmarks. Yet, we envision that studying richer policy classes may results in even better performance. In the end, this work brings together theory and practice: IQ is a theory-consistent manner of implementing an algorithm based on regularized VI in continuous actions settings.

\bibliographystyle{plainnat}
\bibliography{biblio}

\begin{thebibliography}{33}
\providecommand{\natexlab}[1]{#1}
\providecommand{\url}[1]{\texttt{#1}}
\expandafter\ifx\csname urlstyle\endcsname\relax
  \providecommand{\doi}[1]{doi: #1}\else
  \providecommand{\doi}{doi: \begingroup \urlstyle{rm}\Url}\fi

\bibitem[Abdolmaleki et~al.(2018)Abdolmaleki, Springenberg, Tassa, Munos,
  Heess, and Riedmiller]{abdolmaleki2018maximum}
Abbas Abdolmaleki, Jost~Tobias Springenberg, Yuval Tassa, Remi Munos, Nicolas
  Heess, and Martin Riedmiller.
\newblock Maximum a posteriori policy optimisation.
\newblock In \emph{International Conference on learning Representations
  (ICLR)}, 2018.

\bibitem[Akkaya et~al.(2019)Akkaya, Andrychowicz, Chociej, Litwin, McGrew,
  Petron, Paino, Plappert, Powell, Ribas, et~al.]{akkaya2019solving}
Ilge Akkaya, Marcin Andrychowicz, Maciek Chociej, Mateusz Litwin, Bob McGrew,
  Arthur Petron, Alex Paino, Matthias Plappert, Glenn Powell, Raphael Ribas,
  et~al.
\newblock Solving rubik's cube with a robot hand.
\newblock \emph{arXiv preprint arXiv:1910.07113}, 2019.

\bibitem[Azar et~al.(2011)Azar, Ghavamzadeh, Kappen, and
  Munos]{ghavamzadeh2011speedy}
Mohammad~G Azar, Mohammad Ghavamzadeh, Hilbert~J Kappen, and R{\'e}mi Munos.
\newblock {Speedy Q-learning}.
\newblock In \emph{Advances in Neural Information Processing System (NeurIPS)},
  pages 2411--2419, 2011.

\bibitem[Bellemare et~al.(2016)Bellemare, Ostrovski, Guez, Thomas, and
  Munos]{bellemare2016increasing}
Marc~G Bellemare, Georg Ostrovski, Arthur Guez, Philip~S Thomas, and R{\'e}mi
  Munos.
\newblock {Increasing the action gap: New operators for reinforcement
  learning}.
\newblock In \emph{AAAI Conference on Artificial Intelligence (AAAI)}, 2016.

\bibitem[Bertsekas and Tsitsiklis(1996)]{bertsekas1996neuro}
Dimitri~P Bertsekas and John~N Tsitsiklis.
\newblock \emph{Neuro dynamic programming}.
\newblock Athena Scientific Belmont, MA, 1996.

\bibitem[Brockman et~al.(2016)Brockman, Cheung, Pettersson, Schneider,
  Schulman, Tang, and Zaremba]{brockman2016openai}
Greg Brockman, Vicki Cheung, Ludwig Pettersson, Jonas Schneider, John Schulman,
  Jie Tang, and Wojciech Zaremba.
\newblock Openai gym.
\newblock \emph{arXiv preprint arXiv:1606.01540}, 2016.

\bibitem[Dai et~al.(2018)Dai, Shaw, Li, Xiao, He, Liu, Chen, and
  Song]{dai2018sbeed}
Bo~Dai, Albert Shaw, Lihong Li, Lin Xiao, Niao He, Zhen Liu, Jianshu Chen, and
  Le~Song.
\newblock Sbeed: Convergent reinforcement learning with nonlinear function
  approximation.
\newblock In \emph{International Conference on Machine Learning (ICML)}, pages
  1125--1134. PMLR, 2018.

\bibitem[Degrave et~al.(2018)Degrave, Abdolmaleki, Springenberg, Heess, and
  Riedmiller]{degrave2019quinoa}
Jonas Degrave, Abbas Abdolmaleki, Jost~Tobias Springenberg, Nicolas Heess, and
  Martin Riedmiller.
\newblock Quinoa: a q-function you infer normalized over actions.
\newblock \emph{Deep RL Workshop at NeurIPS}, 2018.

\bibitem[Degris et~al.(2012)Degris, White, and Sutton]{degris2012off}
Thomas Degris, Martha White, and Richard~S Sutton.
\newblock Off-policy actor-critic.
\newblock \emph{International Conference on Machine Learning (ICML)}, 2012.

\bibitem[Geist et~al.(2017)Geist, Piot, and Pietquin]{geist2016bellman}
Matthieu Geist, Bilal Piot, and Olivier Pietquin.
\newblock Is the bellman residual a bad proxy?
\newblock \emph{Advances in Neural Information Processing Systems (NeurIPS)},
  2017.

\bibitem[Geist et~al.(2019)Geist, Scherrer, and Pietquin]{geist2019theory}
Matthieu Geist, Bruno Scherrer, and Olivier Pietquin.
\newblock {A Theory of Regularized Markov Decision Processes}.
\newblock In \emph{International Conference on Machine Learning (ICML)}, 2019.

\bibitem[Gu et~al.(2016)Gu, Lillicrap, Sutskever, and Levine]{gu2016continuous}
Shixiang Gu, Timothy Lillicrap, Ilya Sutskever, and Sergey Levine.
\newblock Continuous deep q-learning with model-based acceleration.
\newblock In \emph{International Conference on Machine Learning (ICML)}, 2016.

\bibitem[Haarnoja et~al.(2018{\natexlab{a}})Haarnoja, Zhou, Abbeel, and
  Levine]{haarnoja2018soft}
Tuomas Haarnoja, Aurick Zhou, Pieter Abbeel, and Sergey Levine.
\newblock {Soft actor-critic: Off-policy maximum entropy deep reinforcement
  learning with a stochastic actor}.
\newblock In \emph{International Conference on Machine Learning (ICML)},
  2018{\natexlab{a}}.

\bibitem[Haarnoja et~al.(2018{\natexlab{b}})Haarnoja, Zhou, Hartikainen,
  Tucker, Ha, Tan, Kumar, Zhu, Gupta, Abbeel, et~al.]{haarnoja2018soft2}
Tuomas Haarnoja, Aurick Zhou, Kristian Hartikainen, George Tucker, Sehoon Ha,
  Jie Tan, Vikash Kumar, Henry Zhu, Abhishek Gupta, Pieter Abbeel, et~al.
\newblock Soft actor-critic algorithms and applications.
\newblock \emph{arXiv preprint arXiv:1812.05905}, 2018{\natexlab{b}}.

\bibitem[Hessel et~al.(2018)Hessel, Modayil, Van~Hasselt, Schaul, Ostrovski,
  Dabney, Horgan, Piot, Azar, and Silver]{hessel2018rainbow}
Matteo Hessel, Joseph Modayil, Hado Van~Hasselt, Tom Schaul, Georg Ostrovski,
  Will Dabney, Dan Horgan, Bilal Piot, Mohammad Azar, and David Silver.
\newblock {Rainbow: Combining improvements in deep reinforcement learning}.
\newblock In \emph{AAAI Conference on Artificial Intelligence (AAAI)}, 2018.

\bibitem[Hiriart-Urruty and Lemar{\'e}chal(2004)]{hiriart2004fundamentals}
Jean-Baptiste Hiriart-Urruty and Claude Lemar{\'e}chal.
\newblock \emph{Fundamentals of convex analysis}.
\newblock Springer Science \& Business Media, 2004.

\bibitem[Hoffman et~al.(2020)Hoffman, Shahriari, Aslanides, Barth-Maron,
  Behbahani, Norman, Abdolmaleki, Cassirer, Yang, Baumli, Henderson, Novikov,
  Colmenarejo, Cabi, Gulcehre, Paine, Cowie, Wang, Piot, and
  de~Freitas]{hoffman2020acme}
Matt Hoffman, Bobak Shahriari, John Aslanides, Gabriel Barth-Maron, Feryal
  Behbahani, Tamara Norman, Abbas Abdolmaleki, Albin Cassirer, Fan Yang, Kate
  Baumli, Sarah Henderson, Alex Novikov, Sergio~Gómez Colmenarejo, Serkan
  Cabi, Caglar Gulcehre, Tom~Le Paine, Andrew Cowie, Ziyu Wang, Bilal Piot, and
  Nando de~Freitas.
\newblock Acme: A research framework for distributed reinforcement learning.
\newblock \emph{arXiv preprint arXiv:2006.00979}, 2020.
\newblock URL \url{https://arxiv.org/abs/2006.00979}.

\bibitem[Kingma and Ba(2015)]{kingma2014adam}
Diederik~P Kingma and Jimmy Ba.
\newblock {Adam: A method for stochastic optimization}.
\newblock In \emph{nternational Conference for Learning Representations
  (ICLR)}, 2015.

\bibitem[Metz et~al.(2017)Metz, Ibarz, Jaitly, and Davidson]{metz2017discrete}
Luke Metz, Julian Ibarz, Navdeep Jaitly, and James Davidson.
\newblock Discrete sequential prediction of continuous actions for deep rl.
\newblock \emph{arXiv preprint arXiv:1705.05035}, 2017.

\bibitem[Mnih et~al.(2015)Mnih, Kavukcuoglu, Silver, Rusu, Veness, Bellemare,
  Graves, Riedmiller, Fidjeland, Ostrovski, et~al.]{mnih2015human}
Volodymyr Mnih, Koray Kavukcuoglu, David Silver, Andrei~A Rusu, Joel Veness,
  Marc~G Bellemare, Alex Graves, Martin Riedmiller, Andreas~K Fidjeland, Georg
  Ostrovski, et~al.
\newblock Human-level control through deep reinforcement learning.
\newblock \emph{Nature}, 518\penalty0 (7540):\penalty0 529, 2015.

\bibitem[Nachum et~al.(2017)Nachum, Norouzi, Xu, and
  Schuurmans]{nachum2017bridging}
Ofir Nachum, Mohammad Norouzi, Kelvin Xu, and Dale Schuurmans.
\newblock Bridging the gap between value and policy based reinforcement
  learning.
\newblock \emph{Advances in Neural Information Processing Systems (NeurIPS)},
  2017.

\bibitem[Nachum et~al.(2018)Nachum, Norouzi, Xu, and
  Schuurmans]{nachum2017trust}
Ofir Nachum, Mohammad Norouzi, Kelvin Xu, and Dale Schuurmans.
\newblock Trust-pcl: An off-policy trust region method for continuous control.
\newblock \emph{International Conference on Learning Representations (ICLR)},
  2018.

\bibitem[Rajeswaran et~al.(2017)Rajeswaran, Kumar, Gupta, Vezzani, Schulman,
  Todorov, and Levine]{rajeswaran2017learning}
Aravind Rajeswaran, Vikash Kumar, Abhishek Gupta, Giulia Vezzani, John
  Schulman, Emanuel Todorov, and Sergey Levine.
\newblock Learning complex dexterous manipulation with deep reinforcement
  learning and demonstrations.
\newblock \emph{arXiv preprint arXiv:1709.10087}, 2017.

\bibitem[Rezende and Mohamed(2015)]{rezende2015variational}
Danilo Rezende and Shakir Mohamed.
\newblock Variational inference with normalizing flows.
\newblock In \emph{International Conference on Machine Learning (ICML)}. PMLR,
  2015.

\bibitem[Scherrer et~al.(2015)Scherrer, Ghavamzadeh, Gabillon, Lesner, and
  Geist]{scherrer2015approximate}
Bruno Scherrer, Mohammad Ghavamzadeh, Victor Gabillon, Boris Lesner, and
  Matthieu Geist.
\newblock {Approximate modified policy iteration and its application to the
  game of Tetris}.
\newblock \emph{Journal of Machine Learning Research}, 16:\penalty0 1629--1676,
  2015.

\bibitem[Tang and Agrawal(2020)]{tang2020discretizing}
Yunhao Tang and Shipra Agrawal.
\newblock Discretizing continuous action space for on-policy optimization.
\newblock In \emph{AAAI Conference on Artificial Intelligence (AAAI)},
  volume~34, pages 5981--5988, 2020.

\bibitem[Tavakoli et~al.(2018)Tavakoli, Pardo, and
  Kormushev]{tavakoli2018action}
Arash Tavakoli, Fabio Pardo, and Petar Kormushev.
\newblock Action branching architectures for deep reinforcement learning.
\newblock In \emph{AAAI Conference on Artificial Intelligence (AAAI)},
  volume~32, 2018.

\bibitem[Vieillard et~al.(2020{\natexlab{a}})Vieillard, Kozuno, Scherrer,
  Pietquin, Munos, and Geist]{vieillard2020leverage}
Nino Vieillard, Tadashi Kozuno, Bruno Scherrer, Olivier Pietquin, R{\'e}mi
  Munos, and Matthieu Geist.
\newblock Leverage the average: an analysis of kl regularization in rl.
\newblock In \emph{Advances in Neural Information Processing Systems
  (NeurIPS)}, 2020{\natexlab{a}}.

\bibitem[Vieillard et~al.(2020{\natexlab{b}})Vieillard, Pietquin, and
  Geist]{vieillard2020munchausen}
Nino Vieillard, Olivier Pietquin, and Matthieu Geist.
\newblock Munchausen reinforcement learning.
\newblock \emph{Advances in Neural Information Processing Systems (NeurIPS)},
  2020{\natexlab{b}}.

\bibitem[Wang et~al.(2016)Wang, Schaul, Hessel, Hasselt, Lanctot, and
  Freitas]{wang2016dueling}
Ziyu Wang, Tom Schaul, Matteo Hessel, Hado Hasselt, Marc Lanctot, and Nando
  Freitas.
\newblock Dueling network architectures for deep reinforcement learning.
\newblock In \emph{International conference on machine learning (ICML)}, pages
  1995--2003. PMLR, 2016.

\bibitem[Ward et~al.(2019)Ward, Smofsky, and Bose]{ward2019improving}
Patrick~Nadeem Ward, Ariella Smofsky, and Avishek~Joey Bose.
\newblock Improving exploration in soft-actor-critic with normalizing flows
  policies.
\newblock \emph{arXiv preprint arXiv:1906.02771}, 2019.

\bibitem[Watkins and Dayan(1992)]{watkins1992q}
Christopher~JCH Watkins and Peter Dayan.
\newblock Q-learning.
\newblock \emph{Machine learning}, 8\penalty0 (3-4):\penalty0 279--292, 1992.

\bibitem[Ziebart(2010)]{ziebart2010modeling}
Brian~D Ziebart.
\newblock \emph{{Modeling Purposeful Adaptive Behavior with the Principle of
  Maximum Causal Entropy}}.
\newblock PhD thesis, University of Washington, 2010.

\end{thebibliography}

\clearpage
\onecolumn
\appendix

\paragraph{Contents.}
This Appendix is organized as follows

\begin{itemize}
    \item Appendix~\ref{appx:theory} completes the theoretical claims of the paper
    \begin{itemize}
        \item \ref{subappx:about} details the softmax consistency
        \item  \ref{appx:legendre} gives background on KL and entropy regularization
        \item \ref{appx:adp} proves Theorem~\ref{thm:equivalence} on equivalence between IQ-DP and regularized VI
        \item \ref{appx:error} proves Theorem~\ref{thm:error} on error propagation in IQ-DP
        \item \ref{subappx:munch} details the relation between IQ and Munchausen DQN
    \end{itemize}
    \item Appendix~\ref{appx:experiments} provides details on experiments
    \begin{itemize}
        \item \ref{subappx:info} gives information on our technical setup, regarding software and computing infrastructure
        \item \ref{subappx:pc} details the implementation of our method
        \item \ref{subappx:complete_results} gives a complete view of our experimental results
        \item \ref{subappx:hyperparams} studies the influence of the hyperparameters
    \end{itemize}
\end{itemize}

\section{Analysis}
\label{appx:theory}
This Appendix provides details and proofs on the IQ paramterization.

\paragraph{Reminder on notations.}
Throughout the Appendix, we use the following notations. Recall that we defined the action dot product as, for any $u$ and $v\in\mathbb{R^{\states\times\actions}}$,
\begin{equation}
 \langle u, v\rangle = \big(\sum_{a\in\actions}u(s,a)v(s,a)\big)_s \in \mathbb{R}^\states.
\end{equation}
We also slightly overwrite the $+$ operator. Precisely, for any $Q\in\mathbb{R}^{\states\times\actions}$,  $V\in\mathbb{R^\states}$, we define $Q + V \in \mathbb{R}^{\states\times\actions}$ as
\begin{equation}
    \forall (s,a) \in\states\times\actions, (Q + V)(s,a) = Q(s,a) + V(s).
\end{equation}
Write $\1\in\mathbb{R^{\states\times\actions}}$ the constant function of value $1$. For any $Q\in\mathbb{R}^{\states\times\actions}$, we define the softmax operator as
\begin{equation}
    \sm(Q) = \frac{\exp(Q)}{\langle \1, \exp Q\rangle} \in\mathbb{R^{\states\times\actions}},
\end{equation}
where the fraction is overwritten as the addition operator, that is for any state-action pair $(s,a)$,
\begin{equation}
    \sm(Q)(a|s) = \frac{\exp Q(s,a)}{\sum_{a'\in\actions} \exp Q(s,a')}.
\end{equation}

\subsection{About the softmax consistency}
\label{subappx:about}
First, we provide a detailed explanation of the consistency of the IQ parametrization. In Section~\ref{sec:IQ}, we claim that parametrizing a $Q$-value as $Q = \tau\ln\pi + V$ enforces the relation $\pi = \sm(Q/\tau)$. This relation comes mechanically from the constraint that $\pi$ is a distribution over actions. For the sake of precision, we provide a detailed proof of this claim as formalized in the following lemma.

\begin{lemma}
\label{lemma:softmax}
For any $Q\in\mathbb{R}^{\states\times\actions}$, $\pi\in\Delta_{\actions}^{\states}$, $V\in\mathbb{R^\states}$, we have 
\begin{equation}
\label{eq:softmax-lemmma}
    Q = \tau\ln\pi + V \Leftrightarrow \begin{cases}
                                            \pi = \sm(\frac{Q}{\tau})\\
                                            V = \tau \ln \langle \1, \exp \frac{Q}{\tau}\rangle
                                        \end{cases}.
\end{equation}
\end{lemma}
\begin{proof}
Directly from the left hand side of Eq.~\eqref{eq:softmax-lemmma}, we have
\begin{equation}
    \pi = \exp\frac{Q - V}{\tau}.
\end{equation}
Since $\pi\in\Delta_{\actions}^{\states}$ ($\pi$ is a distribution over the actions), we have 
\begin{align}
\langle \1, \pi \rangle = 1 \Leftrightarrow&  \langle\1, \exp\frac{Q - V}{\tau}\rangle = 1\\
                             \Leftrightarrow&  \left(\exp\frac{-V}{\tau}\right) \langle\1, \exp\frac{Q}{\tau} \rangle = 1 \quad\quad \text{($V$ does not depend on the actions)}\\
                             \Leftrightarrow&  V  = \tau \ln \langle \1, \exp \frac{Q}{\tau}\rangle.
\end{align}
And, for the policy, this gives
\begin{align}
    \pi &= \exp\frac{Q - V}{\tau} = \exp\frac{Q -  \tau \ln \langle \1, \exp \frac{Q}{\tau}\rangle}{\tau} = \frac{\exp(\frac{Q}{\tau})}{\langle \1, \exp \frac{Q}{\tau}\rangle}  = \sm\frac{Q}{\tau}.
\end{align}
\end{proof}

\subsection{Useful properties of KL-entropy-regularized optimization}
\label{appx:legendre}
The following proofs relies on some properties of the KL divergence and of the entropy. Consider the greedy step of MD-VI$((1-\alpha)\tau, \alpha\tau)$, defined in Thm.~\ref{thm:equivalence}
\begin{equation}
\label{eq:greediness}
    \pi_{k+2} = \argmax_{\pi\in\Delta_{\actions}^\states}\langle\pi, r + \gamma P V_k\rangle + (1 - \alpha)\tau \hc(\pi) - \alpha\tau \kl{\pi}{\pi_{k+1}}.
\end{equation}
Since the function $\pi \rightarrow (1 - \alpha)\tau \hc(\pi) - \alpha\tau \kl{\pi}{\pi_{k+1}}$ is concave in $\pi$, this optimization problem can be tackled using properties  of the Legendre-Fenchel transform (see for example~\citet[Chap. E]{hiriart2004fundamentals} for general definition and properties, and~\citet[Appx. A]{vieillard2020leverage} for application to our setting). We quickly state two properties that are of interest for this work in the following Lemma.

\begin{lemma}
\label{lemma:legendre}
Consider the optimization problem of Eq.~\eqref{eq:greediness}. Write $Q_{k+1} = r + \gamma P V_k$, we have that
\begin{equation}
    \pi_{k+2} = \frac{\pi_{k+1}^\alpha \exp\frac{Q_{k+1}}{\alpha}}{\left\langle \1, \pi_{k+1}^\alpha \exp\frac{Q_{k+1}}{\alpha}\right\rangle}.
\end{equation}
We also get a relation between the maximizer and the maximum
\begin{equation}
    \langle\pi_{k+2}, r + \gamma P V_k\rangle + (1 - \alpha)\tau \hc(\pi) - \alpha\tau \kl{\pi}{\pi_{k+1}} = \tau \ln\left\langle \pi_{k+1}^\alpha, \exp \frac{Q_{k+1}}{\tau} \right\rangle.
\end{equation}

\end{lemma}
\begin{proof}
See ~\citet[Appx. A]{vieillard2020leverage}.

\end{proof}
\subsection{Equivalence to MD-VI: proof of Theorem~\ref{thm:equivalence}}
\label{appx:adp}
We turn to the proof of Thm~\ref{thm:equivalence}. This result formalizes an equivalence in the exact case between the IQ-DP scheme and a VI scheme regularized by entropy and KL divergence. Recall that we define the update of IQ-DP at step $k$ as 
\begin{equation}
\label{eq:IQ-DP-appx}
    \tau\ln \pi_{k+2} + V_{k+1} = r + \alpha\tau\ln\pi_{k+1} + \gamma P V_k \quad\quad \text{IQ-DP$(\alpha,\tau)$}.
\end{equation}
Note that we are for now considering the scenario where this update is computed exactly. We will consider errors later, in Thm~\ref{thm:error}. Recall Thm.~\ref{thm:equivalence}.
\begin{recall_thm1}
\label{thm:equivalence-appx}
For any $k\geq1$, let $(\pi_{k+2}, V_{k+1})$ be the solution of IQ-DP at step $k$. We have that
\begin{equation}
\begin{cases}
    \pi_{k+2} = \argmax\langle\pi, r + \gamma P V_k\rangle + (1 - \alpha)\tau \hc(\pi) - \alpha\tau \kl{\pi}{\pi_{k+1}} \\
    V_{k+1} = \langle\pi_{k+2}, r + \gamma P V_k\rangle + (1 - \alpha)\tau \hc(\pi_{k+2}) - \alpha\tau \kl{\pi_{k+2}}{\pi_{k+1}}
\end{cases} 
\end{equation}
so IQ-DP($\alpha, \tau$) produces the same sequence of policies as a value-based version of Mirror Descent VI, MD-VI$(\alpha\tau, (1-\alpha)\tau)$~\citep{vieillard2020leverage}.
\end{recall_thm1}

\begin{proof}
Applying Lemma~\ref{lemma:softmax} to Eq.~\eqref{eq:IQ-DP-appx} gives
\begin{equation}
    \begin{cases}
        \pi_{k+2} = \sm{\frac{r + \alpha\tau\ln\pi_{k+1} + \gamma PV_k}{\tau}}\\
        V_{k+1} = \tau\ln\left\langle 1, \exp \frac{r + \alpha\tau\ln\pi_{k+1} + \gamma PV_k}{\tau} \right\rangle.
    \end{cases}
\end{equation}
For the policy, we have
\begin{align}
    \pi_{k+2} = \frac{\exp\left(\alpha\ln\pi_{k+1}\right) \exp\frac{r +\gamma P V_{k}}{\alpha}}{\left\langle \1, \exp\left(\alpha\ln\pi_{k+1}\right) \exp\frac{r +\gamma P V_{k}}{\alpha}\right\rangle}
              = \frac{\pi_{k+1}^\alpha \exp\frac{r +\gamma P V_{k}}{\alpha}}{\left\langle \1, \pi_{k+1}^\alpha \exp\frac{r +\gamma P V_{k}}{\alpha}\right\rangle},
\end{align}
and as direct consequence of Lemma~\ref{lemma:legendre}
\begin{equation}
    \pi_{k+2} = \argmax\langle\pi, r + \gamma P V_k\rangle + (1 - \alpha)\tau \hc(\pi) - \alpha\tau \kl{\pi}{\pi_{k+1}}.
\end{equation}
For the value, we have:
\begin{equation}
    V_{k+1} = \tau\ln\left\langle 1, \exp(\alpha\ln\pi_{k+1}) \exp \frac{r + \gamma PV_k}{\tau} \right\rangle = \tau\ln\left\langle \pi_{k+1}^\alpha,  \exp \frac{r + \gamma PV_k}{\tau} \right\rangle,
\end{equation}
and again applying Lemma~\ref{lemma:legendre} gives
\begin{equation}
    V_{k+1} = \langle\pi_{k+2}, r + \gamma V_k\rangle + (1 - \alpha)\tau \hc(\pi_{k+2}) - \alpha\tau \kl{\pi_{k+2}}{\pi_{k+1}}.
\end{equation}

\end{proof}

\subsection{Error propagation: proof of Theorem~\ref{thm:error}}
\label{appx:error}
Now we turn to the proof of Thm~\ref{thm:error}. This theorem handles the IQ-DP scheme in the approximate case, when errors are made during the iterations. The considered scheme is
\begin{equation}
\label{eq:IQ-DP-erreappx}
    \tau\ln \pi_{k+2} + V_{k+1} = r + \alpha\tau\ln\pi_{k+1} + \gamma P V_k + \epsilon_{k+1}.
\end{equation}
Recall Thm.~\ref{thm:error}.

\begin{recall_thm2}
\label{thm:error-appx}
Write $\pi_{k+1}$ and $V_k$ the $k^{th}$ update of respectively the target policy and value networks. Consider the error at step $k$, $\epsilon_k\in\mathbb{R}^{\states\times\actions}$, as the difference between the ideal and the actual updates of IQ. Formally, we define the error as, for all $k \geq 1$,
\begin{equation}
    \epsilon_{k+1} = \tau\ln\pi_{k+2} + V_{k+1} - (r + \alpha\tau\ln\pi_{k+1} + \gamma P V_k),
\end{equation}
and the moving average of the errors as
\begin{equation}
    E_k = (1 - \alpha) \sum_{j=1}^{k}\alpha^{k-j}\epsilon_j. 
\end{equation}
We have the following results for two different cases depending on the value of $\alpha$. Note that when $\alpha<1$, we bound the distance to regularized optimal $Q$-value.
\begin{enumerate}
    \item General case: $0 < \alpha <1$ and $\tau> 0$, entropy and KL regularization together:
\begin{equation}
      \|Q_*^{(1-\alpha)\tau} - Q_{\pi_k}^{(1-\alpha)\tau}\|_{\infty}\leq \frac{2}{(1 - \gamma)^2}\left((1-\gamma)\sum_{j=1}^k \gamma^{k-j} \Vert E_j \Vert_\infty \right) + o\left(\frac{1}{k}\right).
\end{equation}
     \item Specific case $\alpha=1$, $\tau > 0$, use of KL regularization alone:
\begin{equation}
     \|Q_* - Q_{\pi_k}\|_{\infty}\leq 
    \frac{2}{1-\gamma} \left\|\frac{1}{k}\sum_{j=1}^k \epsilon_j\right\|_\infty
    + O\left(\frac{1}{k}\right). %
\end{equation}
\end{enumerate}
\end{recall_thm2} 

\begin{proof}
To prove this error propagation result, we first show an extension of Thm.~\ref{thm:equivalence}, that links Approximate IQ-DP with a $Q$-value based version of MD-VI. This new equivalence makes IQ-DP corresponds exactly to a scheme that is extensively analyzed by~\citet{vieillard2020leverage}. Then our result can be derived as a consequence of \citep[Thm 1]{vieillard2020leverage} and~\citep[Thm 2]{vieillard2020leverage}.

Define a (KL-regularized) implicit $Q$-value as
\begin{equation}
\label{eq:q-value_appx}
    Q_{k} =  \tau\ln\pi_{k+1} - \alpha\tau \ln\pi_{k} + V_{k},
\end{equation}
so that now, the IQ-DP update (Eq.~\eqref{eq:IQ-DP-erreappx}) can be written
\begin{equation}
\label{eq:update_err}
    Q_{k+1} = r  + \gamma P V_k + \epsilon_{k+1}.
\end{equation}
We then use same method that for the proof of Thm.~\ref{thm:equivalence}. Specifically, applying Lemma~\ref{lemma:softmax} to the definition of $Q_k$ gives for the policy
\begin{align}
    \pi_{k+1} &= \sm\left(\frac{Q_k + \alpha\tau\ln\pi_{k}}{\tau}\right) \quad\text{(Lemma~\ref{lemma:softmax})}\\
              &= \frac{\pi_{k}^\alpha \exp\frac{Q_{k}}{\alpha}}{\left\langle \1, \pi_{k}^\alpha \exp\frac{Q_{k}}{\alpha}\right\rangle}\\
\Leftrightarrow \pi_{k+1} &= \argmax\langle\pi, Q_k\rangle + (1 - \alpha)\tau \hc(\pi) - \alpha\tau \kl{\pi}{\pi_{k}}. \quad\text{(Lemma~\ref{lemma:legendre})}
\end{align}
For the value, we have from Lemma~\ref{lemma:softmax} on $Q_k$
\begin{align}
    V_{k} = \tau\ln\left\langle 1,  \exp \frac{Q_k + \alpha\tau\ln\pi_{k}}{\tau} \right\rangle = \tau\ln\left\langle \pi_{k}^\alpha,  \exp \frac{Q_k}{\tau} \right\rangle,\\
\end{align}
then, using Lemma~\ref{lemma:legendre}, and the fact that $\pi_{k+1} = \sm\frac{Q_k}{\tau}$, we have
\begin{align}
V_{k} &= \langle\pi_{k+1}, Q_k\rangle + (1 - \alpha)\tau \hc(\pi_{k+1}) - \alpha\tau \kl{\pi_{k+1}}{\pi_{k}}. 
\end{align}
Injecting this in Eq.~\eqref{eq:update_err} gives
\begin{equation}
    Q_{k+1} = r + \gamma P\left(\langle \pi_{k+1}, Q_k \rangle + (1-\alpha)\tau \mathcal{H}(\pi_{k+1}) - \alpha\tau\kl{\pi_{k+1}}{\pi_k}\right).
\end{equation}
Thus, we have proved the following equivalence between DP schemes

\begin{gather}
    \tau \ln \pi_{k+2} + V_{k+1} = r + \alpha\tau\ln\pi_{k+1} +  \gamma P V_k + \epsilon_{k+1}
    \\
    \Updownarrow
    \\
    \begin{cases}
    \label{eq:q-mdvi}
     \pi_{k+1} = \argmax\langle\pi, Q_k\rangle + (1 - \alpha)\tau \hc(\pi) - \alpha\tau \kl{\pi}{\pi_{k}}
      \\
       Q_{k+1} = r + \gamma P\left(\langle \pi_{k+1}, Q_k \rangle + (1-\alpha)\tau \mathcal{H}(\pi_{k+1}) - \alpha\tau\kl{\pi_{k+1}}{\pi_k}\right) + \epsilon_{k+1},
    \end{cases} 
\end{gather}
with 
\begin{equation}
     Q_k = \tau\ln\pi_{k+1} -\alpha\tau\ln\pi_k + V_k.
\end{equation}
The above scheme in Eq.~\eqref{eq:q-mdvi} is exactly the MD-VI scheme studied by~\citet{vieillard2020leverage}, where they define $\beta=\alpha$ and $\lambda=\alpha\tau$. We now use their analysis of MD-VI to apply their result to IQ-DP, building on the equivalence between the schemes. Note that transferring this type of analysis between equivalent formulations of DP schemes is justified because the equivalences exist in terms of \emph{policies}. Indeed, IQ-DP and MD-VI compute different ($Q$)-values, but produce \emph{identical series of policies}. Since \citep[Thm 1]{vieillard2020leverage} and~\citep[Thm. 2]{vieillard2020leverage} bound the distance between the optimal (regularized) $Q$-value and the actual (regularized) $Q$-values of the computed policy, the equivalence in terms of policies is sufficient to apply these theorems to IQ-DP.  Specifically, \citep[Thm 1]{vieillard2020leverage} applied to the formulation of IQ in Eq.~\eqref{eq:q-mdvi} proves point $1$ of Thm.~\ref{thm:error}, that is the case where $\alpha=0$. The second part is proven by applying \citep[Thm 2]{vieillard2020leverage} to this same formulation. 

\end{proof}

\paragraph{Remark on sample complexity.} In the case of a tabular representation (no function approximation needed), and with access to a generative model, we could derive a sample complexity bound from Theorem~\ref{thm:error}. Indeed, we can follow a similar analysis to the one of Speedy Q-learning by~\cite{ghavamzadeh2011speedy} by combining our results with Azuma-Hoeffding concentration inequalities; it  would show that IQ benefits from the same sample complexity as tabular MD-VI~\citep{vieillard2020leverage}, that is $\mathcal{O}\left(\frac{|\states||\actions|}{\epsilon
^{2}(1 - \gamma)^4}\right)$ samples to get an $\epsilon$-optimal policy.

\subsection{IQ and Munchausen DQN}
\label{subappx:munch}
We claim in Section~\ref{sec:analysis} that  IQ is a form of Munchausen algorithm, specifically Munchausen-DQN (M-DQN). Here, we clarify this link. Note that all of the information below is contained in Appx.~\ref{appx:adp} and Appx.~\ref{appx:error}. The point of this section is to re-write it using notations used to defined IQ as a deep RL agent, notations consistent with how M-DQN is defined.

IQ optimizes a policy $\pi_{\theta}$ and a value $V_\phi$ by minimizing a loss $\mathcal{L}_{\text{IQ}}$ (Eq.~\eqref{eq:IQ:loss}). Recall that IQ implicitly defines a $Q$-function as $Q_{\theta, \phi} = \tau\ln\pi_\theta + V_\phi$. Identifying this in $\mathcal{L}_{\text{IQ}}$ makes the connection between Munchausen RL and IQ completely clear. Indeed, the loss can be written as
\begin{equation}
     \hat{\E}\left[\left(r_t + \alpha\tau\ln\pi_{\bar{\theta}}(a_t|s_t) + \underbrace{\gamma V_{\bar{\phi}}(s_{t+1})}_{\tau\ln\sum_{a}\exp \frac{Q_{\bar\theta,\bar\phi}(s_{t+1},a)}{\tau}} - \underbrace{\tau\ln\pi_\theta(a_t|s_t) - V_\phi(s_t)}_{Q_{\theta,\phi}}\right)^2\right],
\end{equation}
and since we have (Lemma~\ref{lemma:legendre}, and using the fact that $\pi_{\bar\theta}=\sm(Q_{\bar\theta,\bar\phi} / \tau)$)
\begin{equation}
    \tau\ln\sum_{a}\exp \frac{Q_{\bar\theta,\bar\phi}(s,a)}{\tau} = \sum_{a}\pi_{\bar\theta}(a | s) \left(Q_{\bar\theta}(s,a) - \tau\ln\pi_{\bar\theta, \bar\phi}(a|s)\right),
\end{equation}
we get that the loss is
\begin{equation}
    \hat{\E}\left[\left(r_t + \alpha\tau\ln\pi_{\bar{\theta}}(a_t|s_t) + \sum_{a}\pi_{\bar\theta}(a|s_{t+1}) \left(Q_{\bar\theta, \bar\phi}(s_{t+1},a) - \tau\ln\pi_{\bar\theta}(a|s_{t+1}) \right) - Q_{\theta,\phi}(s_t,a_t)\right)^2\right],
\end{equation}
which is exactly the Munchausen-DQN loss on $Q_{\theta,\phi}$. Thus, in a mono-dimensional action setting (classic discrete control problems for examle), IQ can really be seen as a re-parameterized version of M-DQN.

\section{Additional material on experiments}
\label{appx:experiments}
This Appendix provides additional detail on experiments, along with complete empirical results.

\subsection{General information on experiments}
\label{subappx:info}
\paragraph{Used assets.} IQ is implemented on the Acme library~\citep{hoffman2020acme}, distributed as open-source code under the Apache License (2.0).

\paragraph{Compute resources.} Experiments were run on TPUv2. One TPU is used for a single run, with one random seed. To produce the main results (without the sweeps over parameters), we computed $780$ single runs. One of this run on a TPUv2 takes from $3$ to $10$ hours depending on the environment (the larger the action space, the longer the run).

\subsection{Details on algorithms}
\label{subappx:pc}

\paragraph{On the relation between $\alpha$ and $\tau$.} The equivalence result of Theorem~\ref{thm:equivalence} explains the role and the relation between $\tau$ and $\alpha$. In particular, it shows that IQ-DP$(\alpha, \tau)$ performs a VI scheme in an entropy-regularized MDP (or in a max-entropy setting) where the temperature is not $\tau$, but $(1-\alpha)\tau$. Indeed, in this framework, the $\alpha$ parameter balances between two forms of regularization: with $\alpha=0$, IQ-DP is only regularized with entropy, but with $\alpha>0$, IQ-DP is regularized with both entropy and KL. Thus, IQ-DP modifies implicitly the intrinsic temperature of the MDP it is optimizing for. To account for this discrepancy, every time we evaluate IQ with $\alpha> 0$ (that is, M-IQ), we report scores using $\tau/(1-\alpha)$, and not $\tau$. For example, on Gym, we used a temperature of $0.01$ for IQ, and thus $0.1$ for M-IQ (since, in our experiments, we took $\alpha=0.9$).

\paragraph{Discretization.} We used IQ with policies that discretize the action space evenly. Here, we provide a precise definition for our discretization method.  Consider a multi-dimensional action space $\actions$ of dimension $d$, each dimension being a bounded interval $[a_{\text{min}}, a_{\text{max}}]$, such that $\actions = [a_{\text{min}}, a_{\text{max}}]^d$. 
We discretize each dimension of the space uniformly  in $n$ values $\delta_j$, for $0\leq j\leq n-1$. The bins values are defined as
\begin{equation}
                                  \delta_0 = a_{\text{min}} + \frac{a_{\text{max}} - a_{\text{min}}}{2n},
\end{equation}
and, for each $j \in\{1,\hdots n-1\},$
\begin{equation}
 \delta_j = \delta_0 + j \frac{a_{\text{max}} - a_{\text{min}}}{n}.
\end{equation}
It effectively defines a discrete action space 
\begin{equation}
    \actions' = \bigtimes_{j=1}^d \actions_j, \quad \text{with} \quad \actions_j = \{\delta_0, \hdots \delta_{n-1}\}.
\end{equation}
 We use $n=11$ in all of our experiments. The values of $d$,  $a_\text{min}$ and $a_\text{max}$ depend on the environments specifications.

\paragraph{Evaluation setting.} We evaluate our algorithms on Mujoco environements from OpenAI Gym and from the Adroit manipulation tasks. On each enviroenment, we track performance for $20$M environment steps. Every $10$k environment steps, we stop learning, and we evaluate our algorithm by reporting the average undiscounted return over $10$ episodes. We use deterministic evaluation, meaning that, at evaluation time, the algorithms interact by choosing the expected value of the policy in one state, not by sampling from this policy (sampling is used during training).

\paragraph{Pseudocode.} We provide a pseudocode of IQ in Algorithm~\ref{algo:iq}. This pseudocode describes a general learning procedure that is followed by all agents. Replacing the IQ loss in Algorithm~\ref{algo:iq} by its residual version will give the pseudocode for PCL, and replacing it by the actor and critic losses of SAC will give the pseudocode for this method.

\begin{algorithm}[tbh]
\caption{Implicit Q-values }
\begin{algorithmic}%
\label{algo:iq}
\REQUIRE $T\in \mathbb{N^*}$ the number of environment steps, $\lambda\in (0,1)$ the update coefficient,   $\gamma \in [0,1)$ the dicount factor,   $\tau \in (0,1)$ the entropy temperature, $\alpha \in [0, 1)$ the implicit KL term, and hyperparameters detailed in Table~\ref{tab:hypers}.
\STATE Initialize $\theta, \phi$ at random
\STATE $\mathcal{B} = \{\}$
\STATE $\bar\theta = \theta$
\STATE $\bar\phi=\phi$
\FOR{$t = 1$ \TO $T$}
    \STATE Collect a transition $b = (s_t, a_t, r_t, s_{t+1})$ from $\pi_\theta$
    \STATE $\mathcal{B} \leftarrow \mathcal{B} \cup \{b\}$
    \STATE On a random batch of transitions $B_{t} \subset \mathcal{B}$, update $(\theta,\phi)$ with one step of SGD on
        \begin{equation}
            \hat{\E}_{(s_t, a_t, r_t, s_{t+1})\sim B_t}\left[\left(r_t + \alpha\tau\ln\pi_{\bar{\theta}}(a_t|s_t) + \gamma V_{\bar{\phi}}(s_{t+1}) - \tau\ln\pi_\theta(a_t|s_t) - V_\phi(s_t)\right)^2\right],
        \end{equation}
    \STATE $\bar\theta \leftarrow \lambda\bar\theta + (1 - \lambda)\theta$
    \STATE $\bar\phi \leftarrow \lambda\bar\phi + (1 - \lambda)\phi$
\ENDFOR
\RETURN $\pi_\theta$
\end{algorithmic}
\end{algorithm}

\paragraph{HyperParameters.}
We Provide the hyperparameters used for our experiments in Tab.~\ref{tab:hypers}. If a parameter is under ``common parameters'', then it was used for all algorithms. We denote $\FC_n$ a fully connected layer with an output of $n$ neurons. Recall that $d$ is the dimension of the action space, and $n$ is the number of bins we discretize each dimension into.

\begin{table}[tbh]
    \centering
    \caption{Parameters used for algorithms and ablations.}
    \begin{tabular}{l r}
    \toprule
    Parameter     & Value \\
    \midrule
    \multicolumn{2}{c}{Common parameters} \\
    \midrule
    $\lambda$ (update coefficient)    & 0.05\\
    $\gamma$ (discount) & 0.99\\
    $|\mathcal{B}|$ (replay buffer size) & $10^6$\\
    $|B_{t}|$ (batch size) & 256 \\
    activations & Relu\\
    optimizer & Adam \\
    learning rate & $3\cdot10^{-4}$ \\
    \midrule
    \multicolumn{2}{c}{IQ specific parameters} \\
    \midrule
    $\tau$ (entropy temperature) & $0.01$ on Gym, $0.001$ on Adroit\\
    $\alpha$ (implicit KL term) & $0.9$ \\
    $n$ (number of bins for the discretization) & 11\\
    $\pi$-network & (input: state) $\FC512-\FC512-\FC nd$\\
    $V$-network structure & (input: state) $\FC512-\FC512-\FC 1$\\
    \midrule
    \multicolumn{2}{c}{(Trust)-PCL specific parameters} \\
    \midrule
     $\tau$ (entropy temperature) & $1\cdot 10^{-4}$ on Gym\\
     $\pi$-network and $V$-network structures & idem as IQ\\
     \midrule
     \multicolumn{2}{c}{SAC specific parameters} \\
    \midrule
     $\pi$-network structure & (input: state) $\FC512-\FC512-\FC 2d$\\
     $Q$-network structure & (input: state and action) $\FC512-\FC512-\FC 1$\\
     learning rate & $3\cdot10^{-4}$ on Gym, $3\cdot10^{-5}$ on Adroit \\
    \bottomrule
    \end{tabular}
    \label{tab:hypers}
\end{table}

\clearpage

\subsection{Additional results}
\label{subappx:results}
This Appendix section provides complete description of the scores and ablations in~\ref{subappx:complete_results}, along with a study of some hyperparameters in~\ref{subappx:hyperparams}.

\subsubsection{Complete scores and ablations}
\label{subappx:complete_results}
\begin{figure}
    \centering
    \includegraphics[width=.49\linewidth]{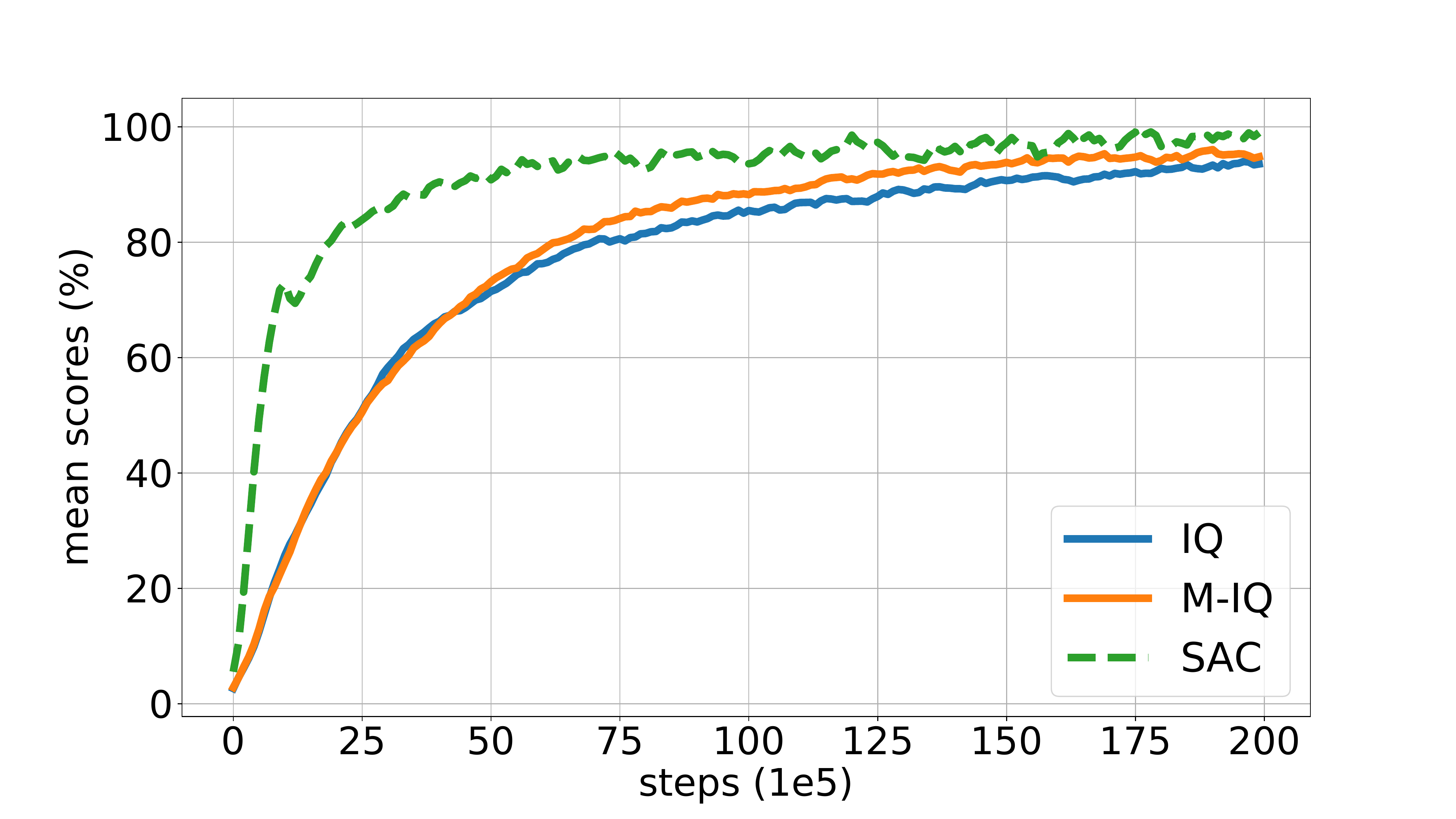}
    \includegraphics[width=.49\linewidth]{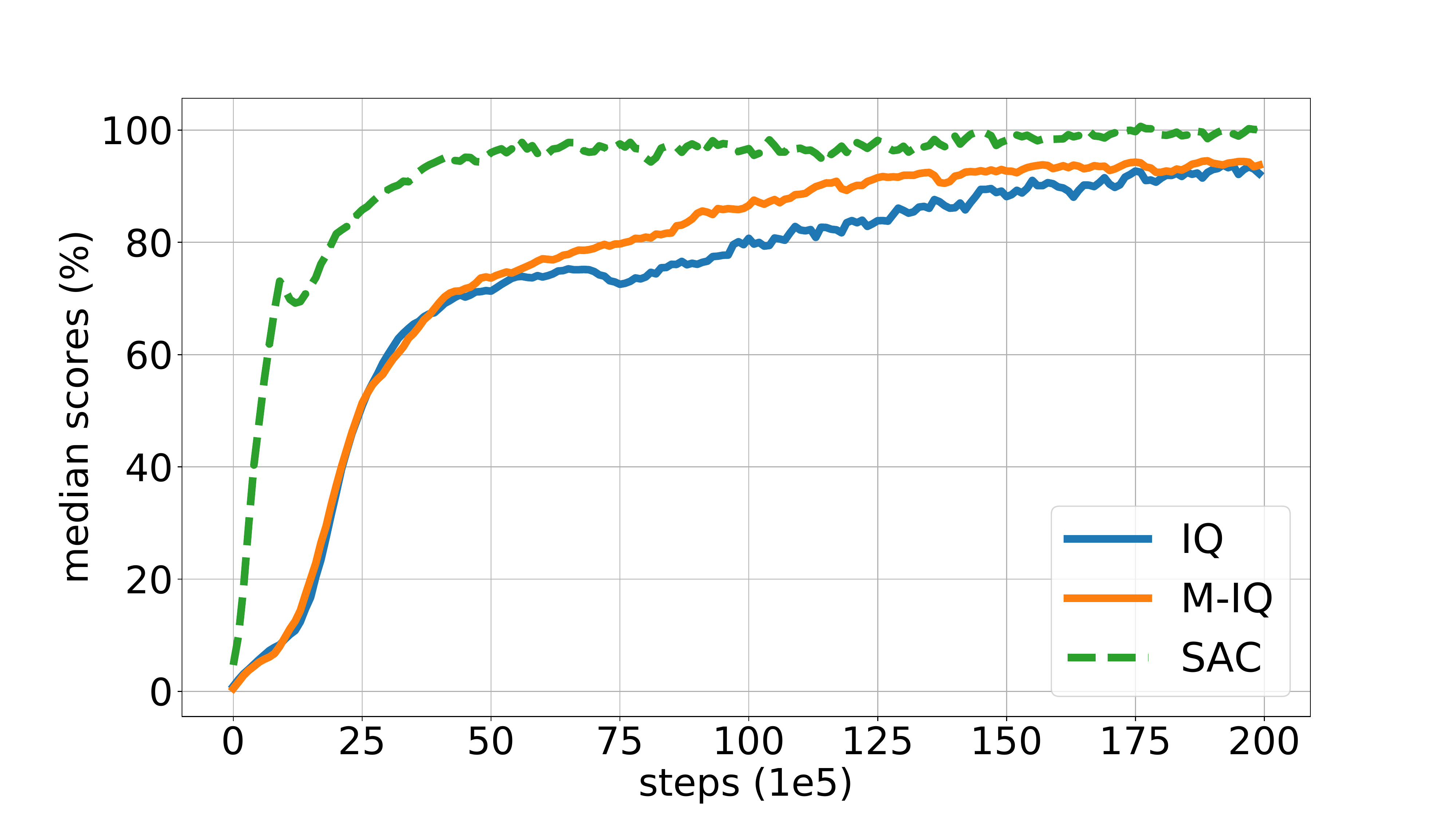}
    \caption{SAC-normalized aggregated scores on Gym environments. \emph{Best parameters}: IQ uses a different learning rate for Humanoid-v2 and Walker2d-v2. \textbf{Left}: Mean scores. \textbf{Right:} Median scores.}
    \label{fig:mujoco_all_appx}
\end{figure}

\begin{figure}
    \centering
    \includegraphics[width=.49\linewidth]{figures/nlc_adroit_mean_20s.pdf}
    \includegraphics[width=.49\linewidth]{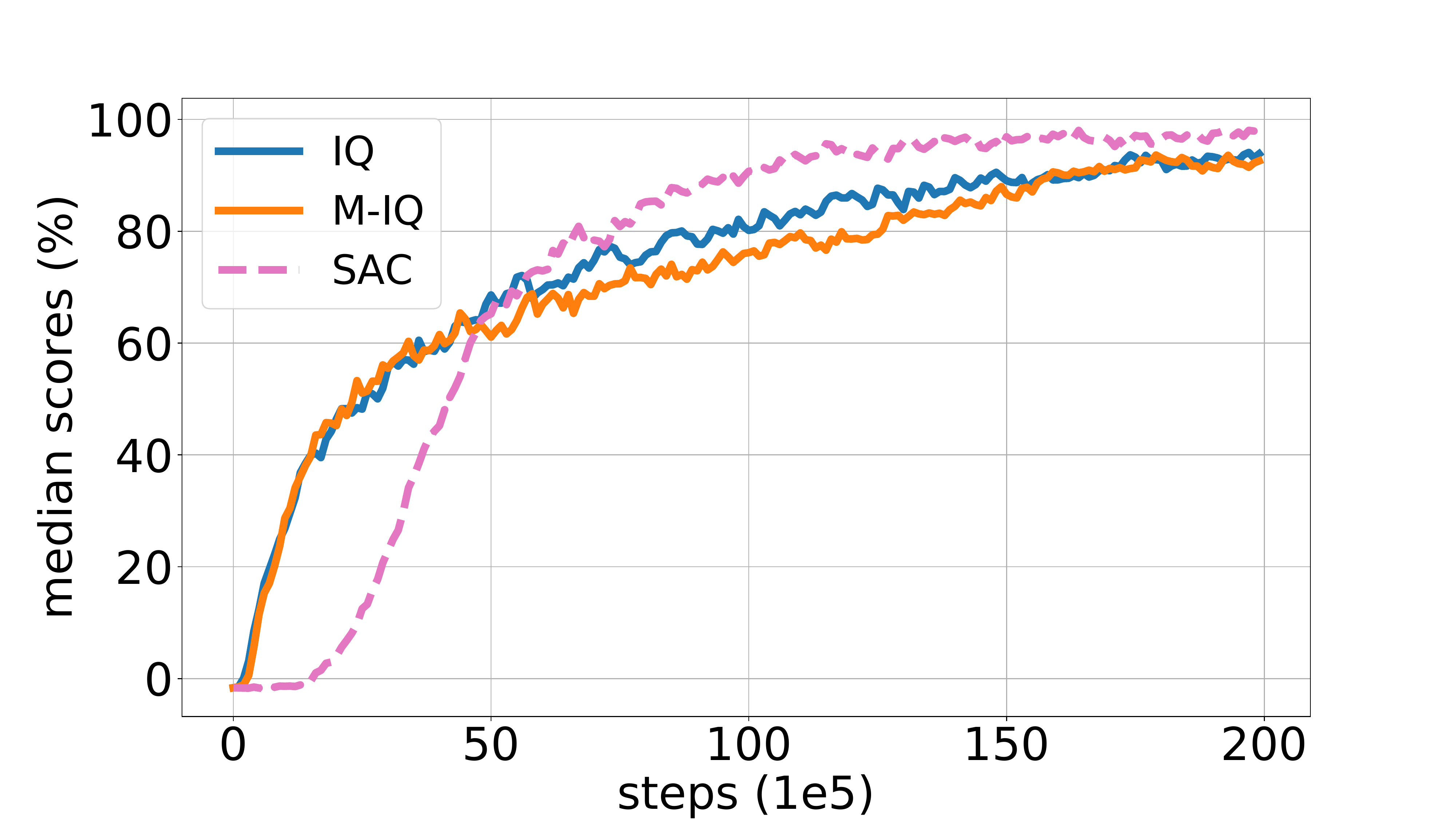}
    \caption{SAC-Normalized aggregated scores on Adroit. \textbf{Left}: Mean scores. \textbf{Right:} Median scores.}
    \label{fig:adroit_all_appx}
\end{figure}

Here, we provide detailed results in a more readable fashion in Fig.~\ref{fig:mujoco_all_appx} and~\ref{fig:adroit_all_appx}. We also give more details on the aggregated scores (median scores, explanations on the hyperparmeter selection), and provide a more extensive analysis of our experiments.

\begin{figure}
    \centering
    \includegraphics[width=\linewidth]{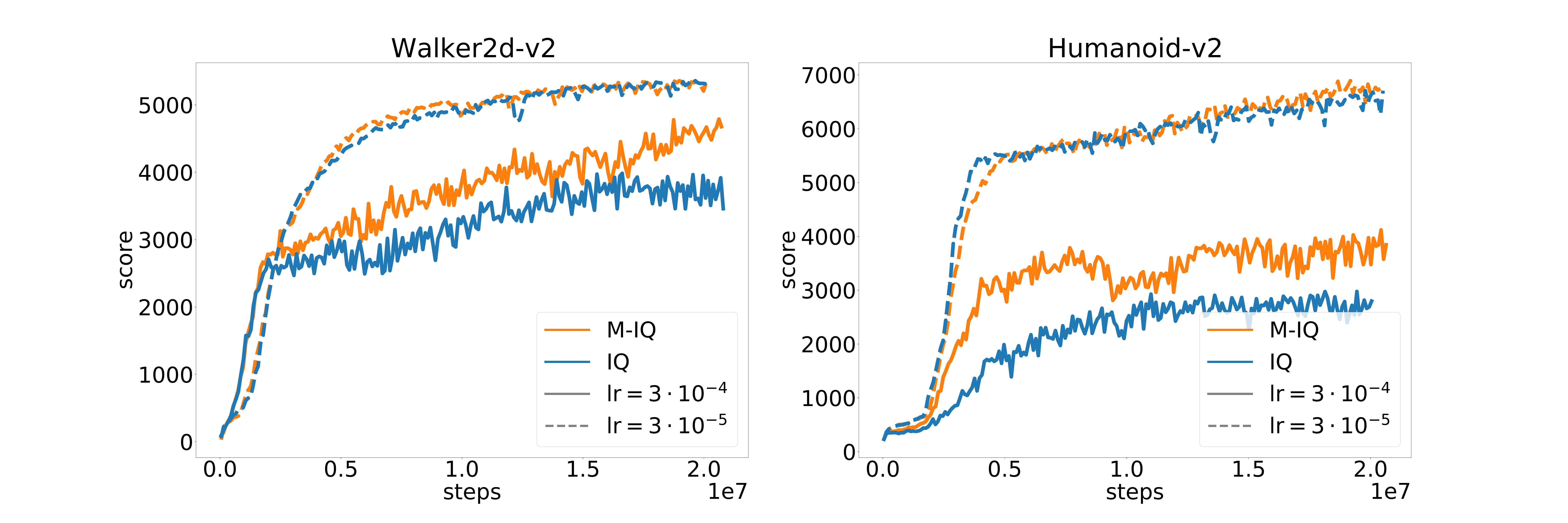}
    \caption{Comparison of two value of the learning rate (lr) on IQ and M-IQ, on two environments. Each line corresponds to the average score over $20$ seeds.}
    \label{fig:lr_comparison}
\end{figure}

\begin{figure}
    \centering
    \includegraphics[width=.49\linewidth]{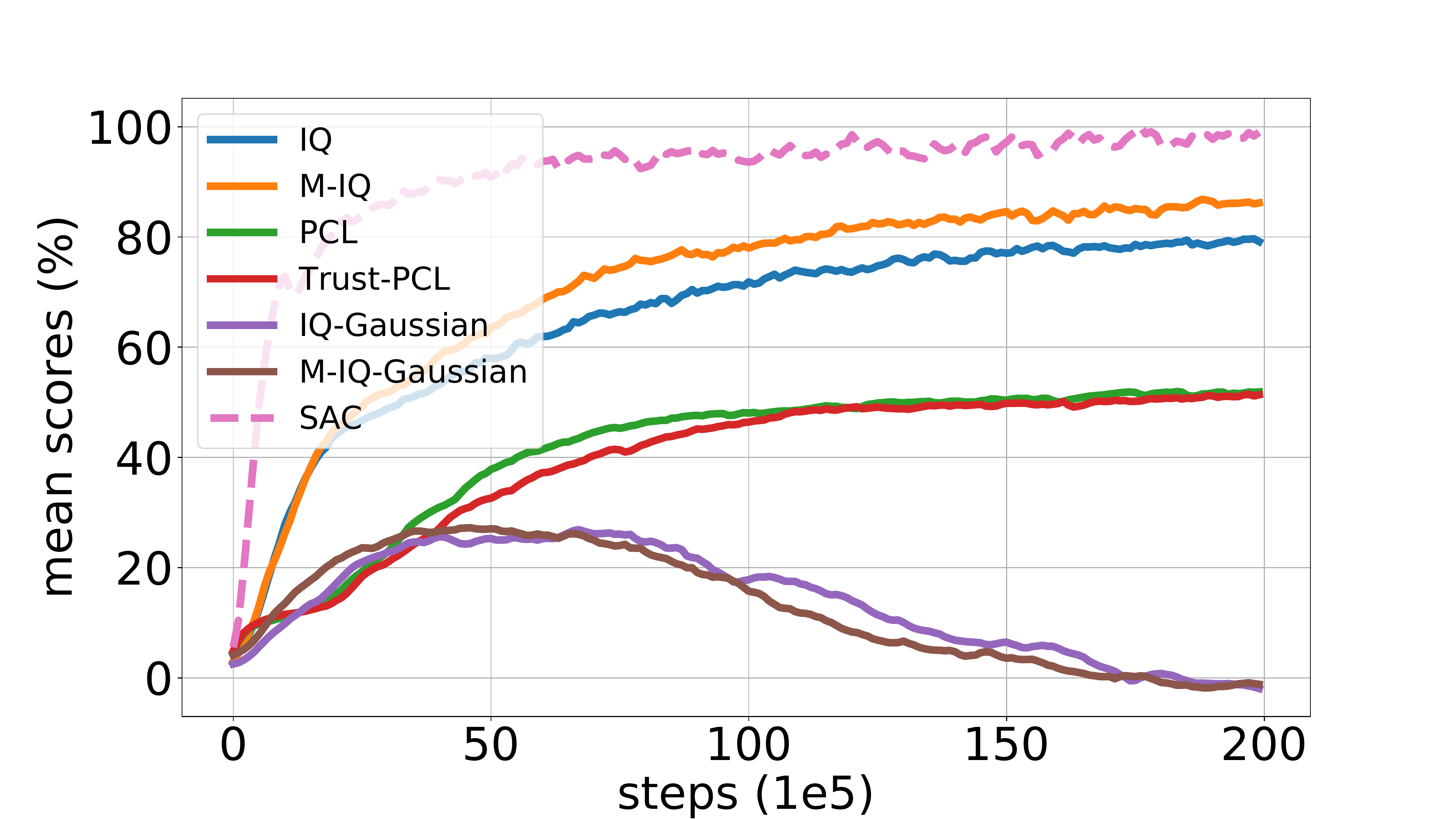}
    \includegraphics[width=.49\linewidth]{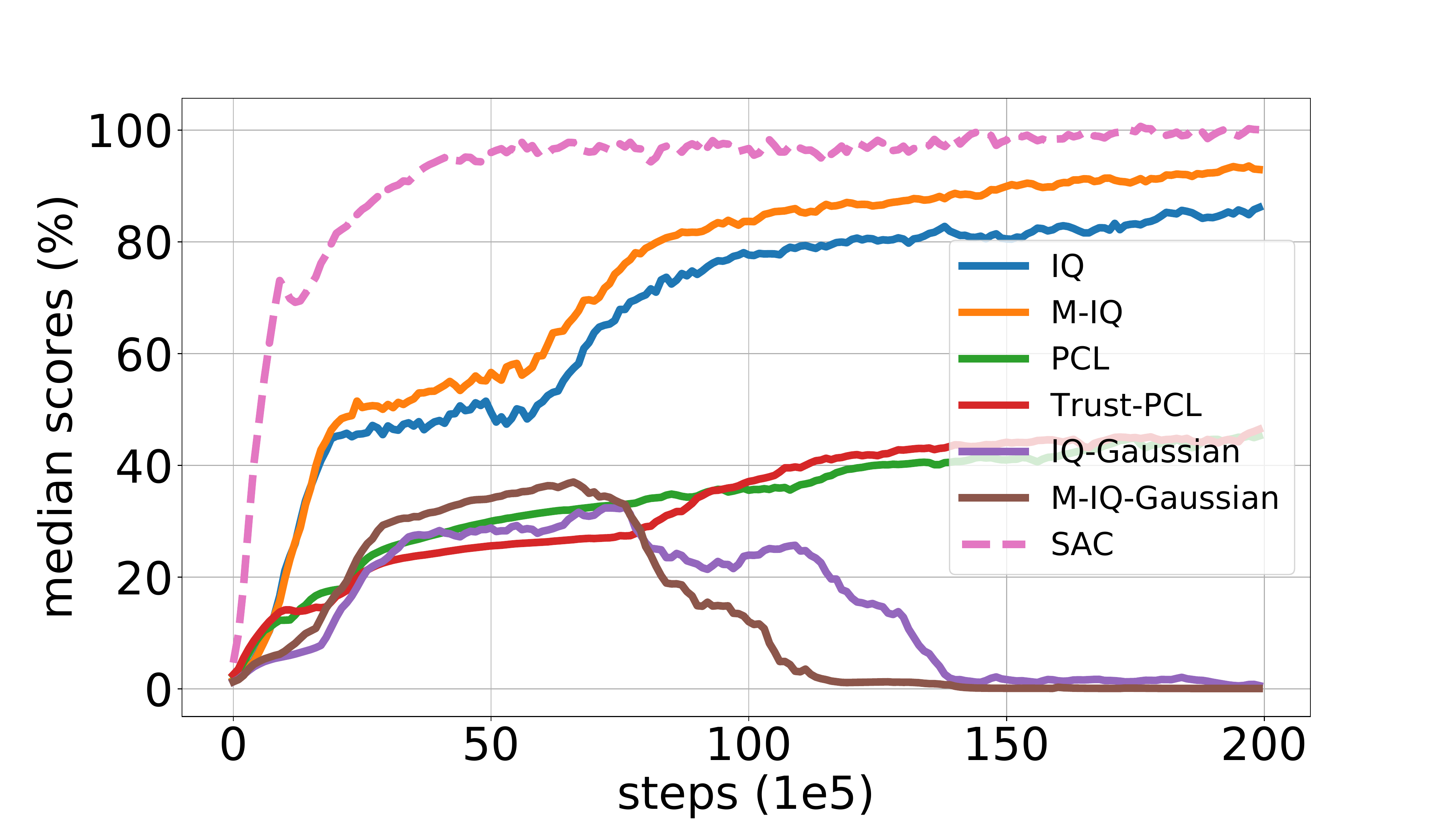}
    \caption{SAC-normalized ablation scores on Gym. \emph{Unique parameters}: all algorithms use a single set of hyperparmeters for the $5$ environments.}
    \label{fig:mujoco_all_appx_2}
\end{figure}

\paragraph{Aggregated scores.} Aggregated scores on Gym and Adroit environments are reported in Figures~\ref{fig:mujoco_all_appx_2} and~\ref{fig:adroit_all_appx}. We see from these results that (M)-IQ reaches performance close to SAC, but is still slightly below. On Gym, IQ is slower (as in less sample efficient) than SAC, but is faster on Adroit. One important thing to notice is that we compare to the best version of SAC we found, that uses an adaptive temperature, while we use the same fixed temperature for all of the environments. Experimentally, it appears that IQ is more sensitive to hyperparameters than SAC\footnote{However, we note that SAC, when used with a fixed temperature, is sensitive to the choice of this parameter. It appears that IQ is less sensitive to this (we use the same temperature for all tasks), but a bit more to the learning rate. We think this may be alleviated by adopting other policy representations, such as normalizing flows, or by designing adaptive schemes. We left this for future works.}. Notably, on Humanoid-v2 and Walker2d-v2, we show in Figure~\ref{fig:lr_comparison} that IQ can reach a much more competitive score by using a specific hyperparameter per environment. To reflect this, we use a learning rate of $3\cdot 10^{-5}$ instead of $3\cdot 10^{-4}$ for Humanoid and Walker, while still comparing to the best version of SAC we found (we also tested SAC with those parameters variations, but it did not improve the performance). This is reported as \textit{best parameters} results in Fig.~\ref{fig:mujoco_all_appx}; and for completeness, we report score using \emph{unique hyperparameters} in Fig.~\ref{fig:mujoco_all_appx_2}. On Adroit, we did not observe such variations, and thus we report scores with a single set of hyperparameters.

\begin{figure}
    \centering
   \includegraphics[width=\linewidth]{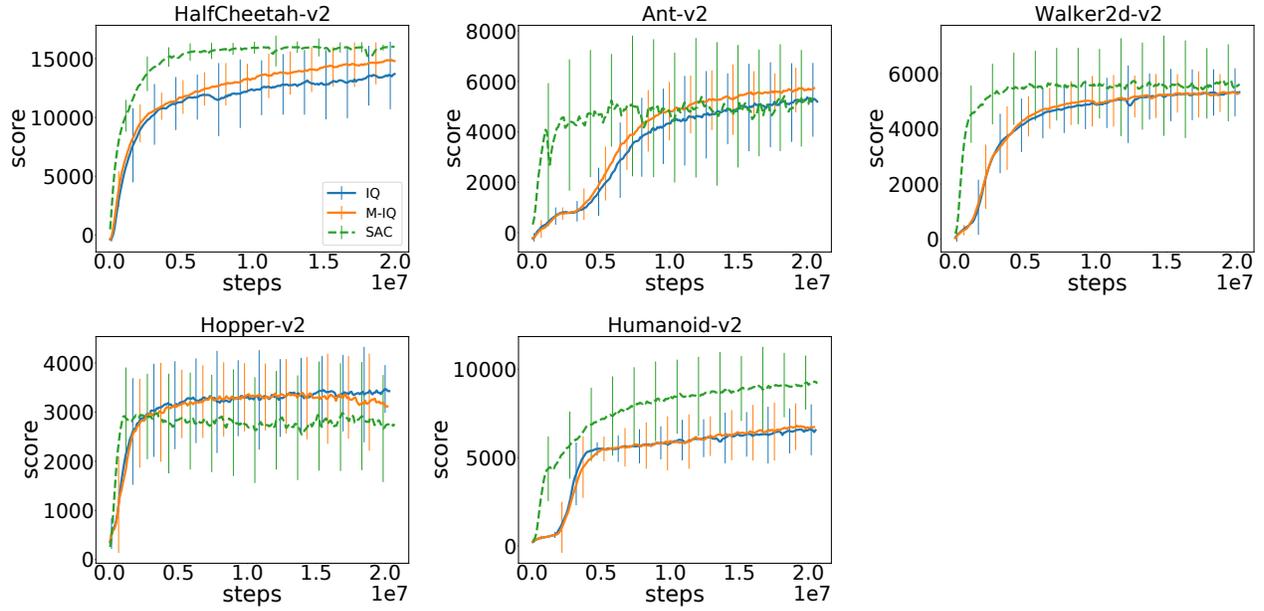}
    \caption{All individual scores on Gym. The vertical bars denote the empirical standard deviation over $20$ seeds. \emph{Best parameters}: IQ uses a different learning rate for Humanoid-v2 and Walker2d-v2.}
    \label{fig:mujoco_envs_best}
\end{figure}

\begin{figure}
    \centering
   \includegraphics[width=\linewidth]{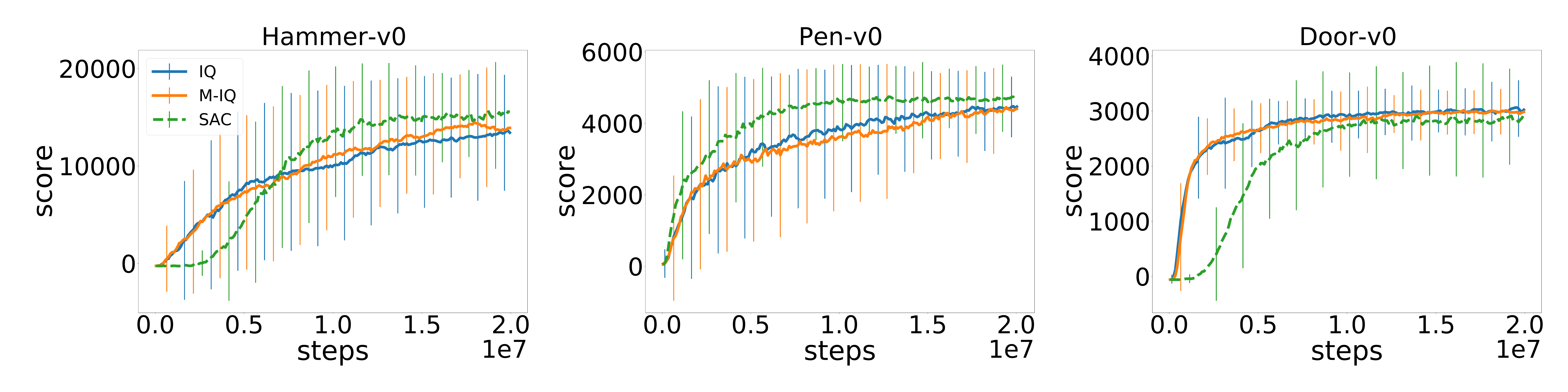}
    \caption{All scores on Adroit. The vertical bars denote the empirical standard deviation over $20$ seeds.}
    \label{fig:adroit_envs}
\end{figure}

\paragraph{Detailed scores.} We report detailed scores on Gym environments in Figure~\ref{fig:mujoco_envs_best} and on Adroit in Figure~\ref{fig:adroit_envs}. On this figures, the thick line corresponds to the average over $20$ seeds, and the error bars represents $+/-$ the empirical standard deviation over those seeds. It shows how the performance varies across environments. Specifically, IQ outperforms SAC on $3$ of the $8$ considered environments: Ant, Hopper, and Door. On the $5$ others, SAC performs better, but IQ and M-IQ still reach a reasonable performance. Moreover, on almost all environments, all mean performances are within the same confidence interval (Humanoid, and to a less extent HalfCheetah, being notable exceptions, in favor of SAC).

\paragraph{Ablations.} We report ablation scores, averaged over $20$ seeds, comparing IQ, M-IQ, PCL, Trust-PCL and (M)-IQ-Gaussian in Fig.~\ref{fig:mujoco_all_appx_2}. PCL (and Trust-PCL) are obtained by replacing $V_{\bar\phi}$ by $V_\phi$ in $\mathcal{L}_\text{IQ}$. (M)-IQ Gaussian is obtained by parametrizing the policy as a diagonal normal distribution over the action. Results confirm that the fixed-point approach is beneficial wrt the residual one in this case, since IQ outperforms PCL  and trust-PCL by a clear margin in median and mean\footnote{We note that PCL was introduced only for discrete actions, and that Trust-PCL was proposed with a Gaussian policy, on a much shorter time frame for training, and with many additional tricks.}. They also validate that the Gaussian parametrization is too limiting in IQ, since IQ-Gaussian and M-IQ-Gaussian both totally collapse after $6$ to $7$ millions environment steps. Additionally, this also highlights that the time frame usually considered on some of this environments ($3$M, or even $1$M steps) may be too short to actually observe the asymptotic behavior of many agents. Here, this time frame would have prevented us to observe this collapsing phenomenon on the Gaussian IQ. 

\subsubsection{Influence of hyperparameters}
\label{subappx:hyperparams}
In addition to the "deep" hyperparameters that control the function approximation (layer sizes, activations, learning rate, ..), and that we take from our baselines, (M)-IQ relies on three key hyperparameters. Those are the Munchausen coefficient $\alpha$, the temperature $\tau$, and the number of bins $n$. The influence of $\alpha$ is represented by the difference between M-IQ and IQ. Here, we study the influence of the two others hyperparameters, $\tau$ and $n$.

\begin{figure}
    \centering
   \includegraphics[width=\linewidth]{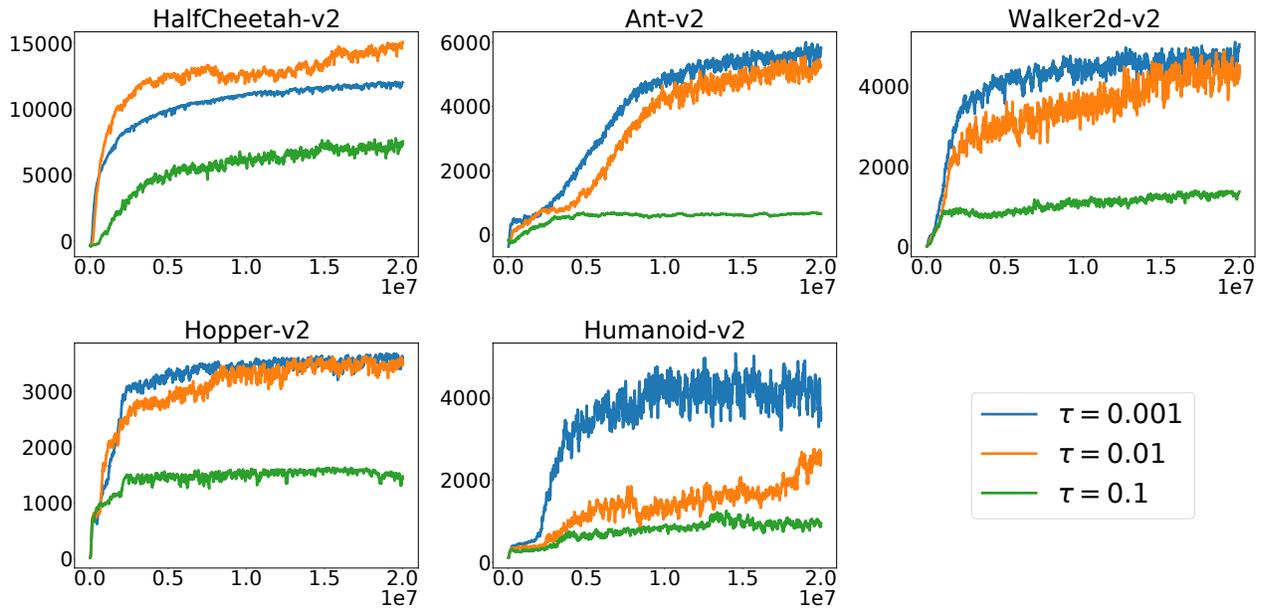}
    \caption{IQ ($\alpha=0$) with several values for $\tau$ on all Gym tasks. Each line is the average over $10$ seeds.}
    \label{fig:mujoco_temp_appx}
\end{figure}

\begin{figure}
    \centering
   \includegraphics[width=\linewidth]{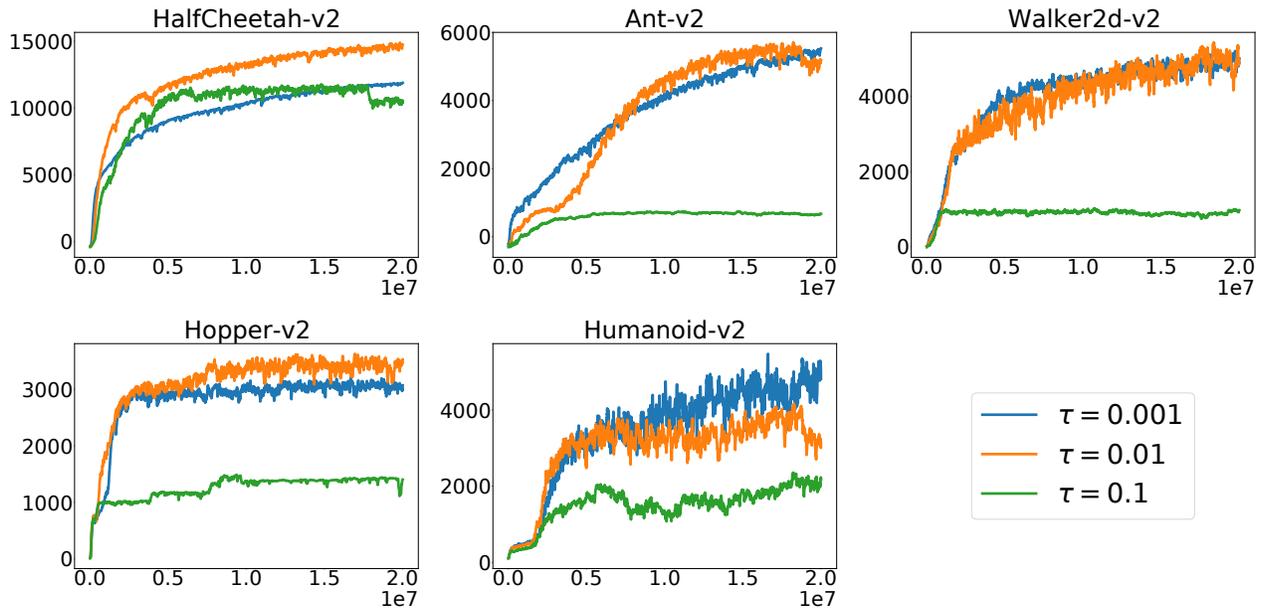}
    \caption{M-IQ ($\alpha=0.9$) with several values for $\tau$ on all Gym tasks. Each line is the average over $10$ seeds.}
    \label{fig:mujoco_temp_m_appx}
\end{figure}

\paragraph{Influence of the temperature ($\tau$).} We provide results for $3$ values of $\tau$ and $2$ values of $\alpha$. We report scores for IQ with $\alpha=0$ with different values of $\tau$ in Fig.~\ref{fig:mujoco_temp_appx} and the same experiments on M-IQ (with $\alpha=0.9$) in Fig.~\ref{fig:mujoco_temp_m_appx}. We can draw two observations from these results. \textbf{(1)} They show that $\tau$ needs to be selected carefully: it helps learning, but too high values can be highly detrimental to the performance. The temperature also depends on the task. While we could find a value that give reasonable results on all the considered environment, this means that adapting the automatic scheduling of the temperature used in SAC could be a promising direction for IQ. \textbf{(2)} The other observation is that clearly, $\tau$ has a much stronger influence on IQ than $\alpha$. This is a key empirical difference regarding the performance of M-DQN~\citep{vieillard2020munchausen}, that has the same parameters, but is evaluated on discrete actions settings. In these settings, the $\alpha$ parameters is shown to have a crucial importance in terms of empirical results: M-DQN with $\alpha=0.9$ largely outperforms M-DQN with $\alpha=0$ on the Atari benchmark. While this term still has effect in IQ on some tasks, it is empirically less useful, even though it is never detrimental.  We conjecture that this discrepancy comes from the inductive bias we introduce with the multicategorical policy, that could lessen the effect of the Munchausen trick. 

\begin{figure}
    \centering
    \includegraphics[width=\linewidth]{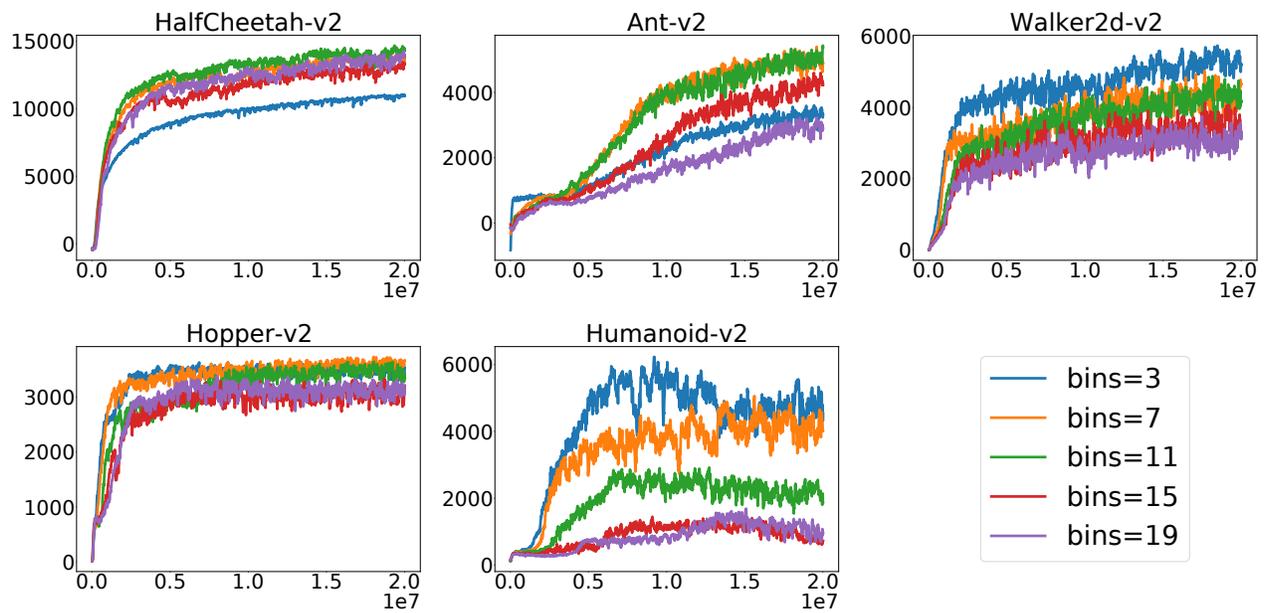}
    \caption{IQ with several number of bins on all Gym environments. Each line is the average over $10$ seeds.}
    \label{fig:mujoco_bins_appx}
\end{figure}

\paragraph{Influence of the number of bins.} In Fig.~\ref{fig:mujoco_bins_appx}, we show how the number of bins impact the performance of IQ on the Gym environments. The results indicate that this parameter is fairly easy to tune, as IQ is not too sensitive to its variations. A remarkable observation is that with only $3$ bins, IQ can achieve reasonable performance on complex environments (see the results on Humanoid for example). This turns out to be a beneficial feature of the multicategorical policies. Indeed, the limiting factor of this parametrization, is that wee need a potentially large number of bins in total, if the dimensions of the environment grows: the output size of the policy network is $d * n$, with $d$ the dimension of the action space and $n$ the number of bins. If  we can keep the number of bins close to a minimal value (as it seems to be the case here), this is promising for potential use of IQ on larger action spaces.

\end{document}